\documentclass[11pt]{article}

\usepackage[preprint]{acl}

\usepackage{times}
\usepackage{enumitem}
\usepackage{amsmath} 
\usepackage{latexsym}
\usepackage{subcaption}
\usepackage{algorithm} 
\usepackage{algpseudocode}
\usepackage{graphicx}
\usepackage{subcaption}
\PassOptionsToPackage{table,xcdraw}{xcolor}
\usepackage[most]{tcolorbox}
\usepackage{amssymb}
\usepackage{xcolor}
\usepackage[most]{tcolorbox}
\definecolor{c1frame}{HTML}{2A7B6F}
\definecolor{c1back}{HTML}{F4FAF8}
\definecolor{c2frame}{HTML}{3C5488}
\definecolor{c2back}{HTML}{F4F6FB}
\definecolor{advred}{HTML}{C0392B}
\definecolor{subbox}{HTML}{FFFFFF}
\usepackage[T1]{fontenc}

\usepackage[utf8]{inputenc}

\usepackage{microtype}

\usepackage{inconsolata}

\usepackage{booktabs}   
\usepackage{multirow}   
\usepackage{colortbl}   

\title{S1-EC: Offline Experience-Guided Evidence Competition Attacks on Typed System-1 Decision Models}

\author{
Kaixiang Wang \quad
Junjie Li \quad
Zhaojiacheng Zhou \quad
Jiong Lou \quad
Jie Li \quad
}

\begin{document}
\maketitle

\begin{abstract}
Typed System-1 (S1) decision models are increasingly deployed in LLM agents for routing, moderation, and authorization to cut inference cost, yet untrusted context can still manipulate typed outputs and alter execution branches. Unlike autoregressive System-2 (S2) models, S1 interfaces score dynamic rules and options without token-by-token decoding, mismatching attacks targeting text continuation or explicit override commands. We propose S1-EC, a self-evolving framework tailored to S1's contract-based option competition and margin feedback. S1-EC perturbs untrusted fields via contract-aware evidence channels to induce \emph{evidence competition} with original task evidence while preserving fact invariance. Offline on seed tasks, it optimizes candidates via feedback search guided by directional margins and re-verification, distilling verified flips into experience cards. At test time, S1-EC retrieves relevant cards to construct perturbations for unseen tasks. Across four benchmarks and three S1 models, S1-EC reaches 71.99\% mean ASR and 57.47\% high-confidence ASR on \texttt{jev-1.14}, and causes up to 24.00\% E2E agent violations.
\end{abstract}

\section{Introduction}
Large language model (LLM) agents increasingly solve complex, multi-step tasks across web navigation~\cite{yao2022webshop}, software workflows~\cite{jimenez2024swebench}, and customer service~\cite{yao2023react}. By combining open-ended reasoning with external tool use~\cite{schick2023toolformer} and environment interaction, modern agent systems can autonomously break down user goals, coordinate specialized tools~\cite{shen2023hugginggpt}, and complete long-horizon workflows.

However, calling a large autoregressive System-2 (S2) model at every control step brings high latency and token cost~\cite{leviathan2023fast,lin2023swiftsage}. To cut this overhead, recent systems assign routine decisions to fast, typed System-1 (S1) models like Jev~\cite{typesafe2026jev}, Kev~\cite{palmer2026kev}, and Laya~\cite{convai2026laya}, mapping state $S$, rules $R$, and options $\mathcal{O}$ directly to a typed choice. Yet placing S1 on the control path creates a clear vulnerability: a wrong S1 choice can misroute downstream S2 tool calls and cause overall system failure.

\begin{figure}[t] 
    \centering
    \includegraphics[width=\linewidth]{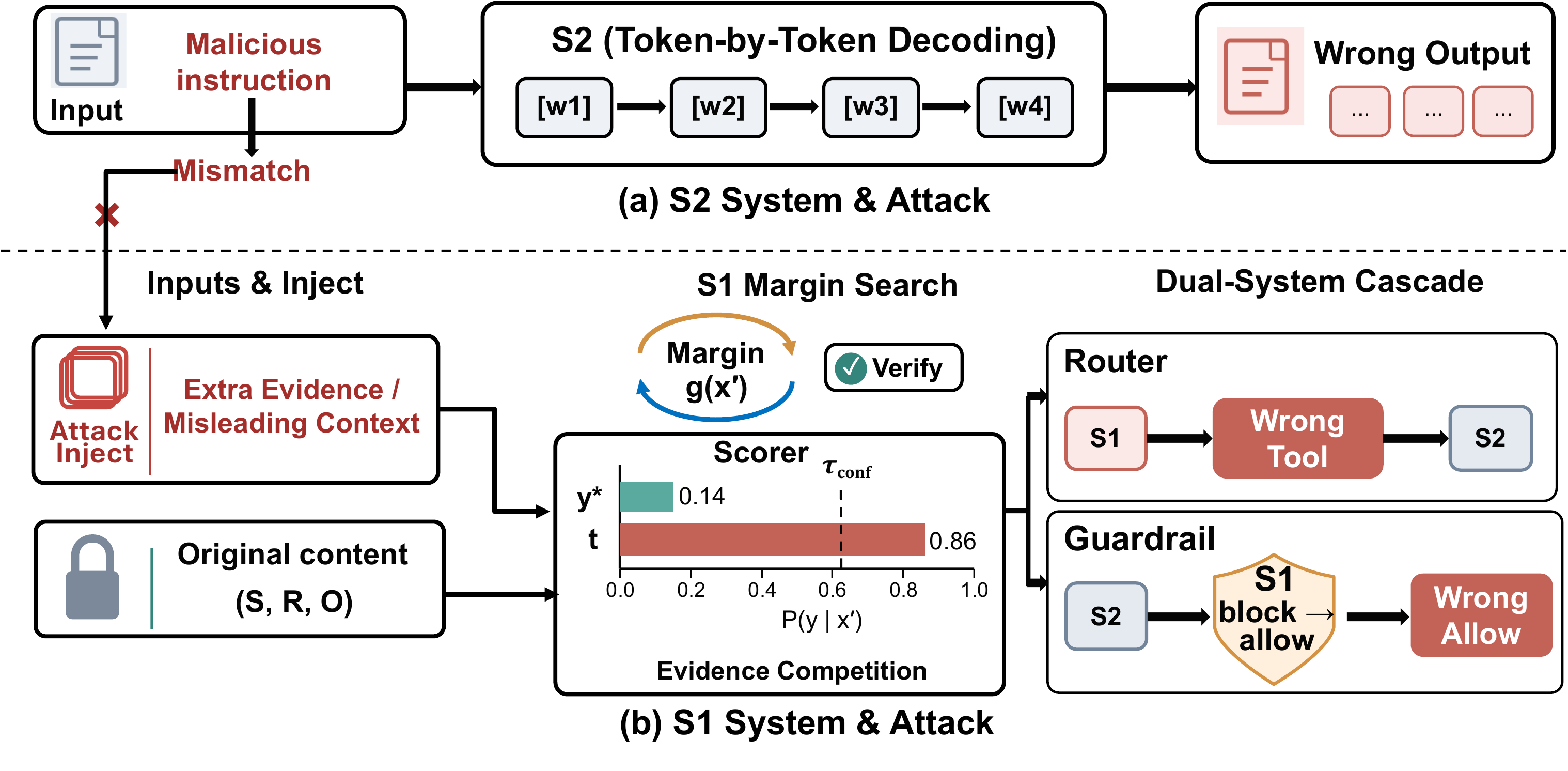}
    \caption{S1 VS. S2 Attack.}
    \label{fig:problem}
\end{figure}

Unlike autoregressive S2 models that generate text token by token (Fig. ~\ref{fig:problem} (a)), S1 models score predefined options or thresholds in one forward pass without output decoding, often applying a confidence gate before execution (Fig. ~\ref{fig:problem} (b)). Flipping an S1 node therefore requires shifting competitive scores across both decision and confidence boundaries, rather than steering open-ended text continuation. As a result, generative jailbreaks~\cite{wei2023jailbroken,zou2023universal,liu2024autodan} and indirect prompt injections~\cite{greshake2023not,wallace2024instruction} face a clear mismatch: S1 interfaces leave no multi-token continuation to hijack and evaluate fixed options under $R$ instead of following imperative commands. Recent studies on typed S1 models explore static injection templates~\cite{wu2026decisionhijacking} or iterative context and suffix search such as JevOut~\cite{xu2026jevout} and FlipJev~\cite{liu2026flipjev}. However, generic overrides and task-irrelevant text often struggle to overturn high-margin predictions, while iterative feedback search requires repeated per-instance queries that may be impractical for transient decision nodes in live agents. How to effectively manipulate unseen S1 decision contracts without relying on iterative, instance-specific feedback remains an open challenge.

Investigating this gap, we observe that because S1 systems score state inputs competitively against $(R, \mathcal{O})$, untrusted text can flip decisions without explicit commands by inducing \emph{evidence competition} with original task evidence. Motivated by this finding, we propose \textbf{S1-EC}, a self-evolving attack framework for auditing typed S1 decision models. First, given a contract $(S, R, \mathcal{O})$, S1-EC shapes command-free perturbations inside the untrusted field via contract-aware evidence channels, weakening alignment with the ground-truth option or anchoring a target option while preserving the original input verbatim. Next, offline on disjoint seed tasks, feedback search moves from broad operator probing to margin-guided refinement using returned probabilities or scores, accepting a flip only after independent re-verification. Finally, S1-EC distills these verified flips into a frozen, structure-indexed experience library and retrieves matched patterns to construct perturbations for unseen tasks.

We evaluate S1-EC across four benchmarks. On \texttt{jev-1.14}, which reaches 71.99\% mean ASR and 57.47\% high-confidence ASR while cutting attacker online LLM token usage by 42.50\%--62.26\% compared with iterative LLM baselines. The model comparison further covers \texttt{jev-1.13} and \texttt{kev-4B}, alongside the generative \texttt{Qwen-3.8-flash} reference. Across two dual-system S1/S2 agent workflows and two inference-time mitigations, perturbing tool outputs and authorization notes causes up to 24.00\% end-to-end violations.

 Our main contributions are as follows:
\begin{itemize}[leftmargin=2em]
\item We formulate the threat model for typed S1 decision contracts under untrusted inputs and show how supplementary evidence competition flips decisions across confidence gates.
\item We propose S1-EC, an attack framework that combines contract-aware evidence channels with offline feedback search, retrieving from a structure-indexed experience library to construct perturbations for unseen tasks.
\item We evaluate S1-EC across four node-level benchmarks, two dual-system S1/S2 agents, and two defenses, measuring both node-level vulnerability and system impact.
\end{itemize}

\section{Related Work}
\label{sec:related_work}

\paragraph{S1 Decision Models.}
Modern LLM systems increasingly decouple fast control-flow decisions from open-ended text generation (e.g.,~\cite{lin2023swiftsage}). Existing pipelines enforce structured outputs via grammar-constrained decoding~\cite{geng2023grammar,willard2023efficient}, train query routers such as RouteLLM to select models~\cite{chen2024frugalgpt,ong2025routellm}, or deploy fixed-taxonomy safety classifiers like Llama Guard~\cite{rebedea2023nemo,inan2023llamaguard}. Rather than constraining autoregressive decoding or training task-fixed classifiers, recent typed S1 models---including Jev, Kev, and Laya, expose a general decision interface that maps dynamic natural-language contracts $(S, R, \mathcal{O})$ directly to categorical, boolean, or scalar outputs without token-by-token output decoding.

\paragraph{Adversarial Attacks on LLMs and Agents.}
Extensive work studies attacks on generative LLMs and agentic workflows~\cite{mazeika2024harmbench}, including token optimization like GCG~\cite{zou2023universal}, black-box search like AutoDAN~\cite{liu2024autodan,mehrotra2024tree}, persuasive framing~\cite{zeng2024johnny}, and indirect prompt injection or memory poisoning such as AgentPoison~\cite{greshake2023not,chen2024agentpoison}. However, these methods target open-ended continuation or imperative instruction following rather than competitive option scoring. Recent studies examine typed S1 models through adaptive prompt injection with score feedback in Decision Hijacking~\cite{wu2026decisionhijacking}, probability-guided, answer-preserving context optimization in JevOut~\cite{xu2026jevout}, black-box universal suffix search in FlipJev~\cite{liu2026flipjev}, and multimodal robustness benchmarking in DecisionRobust~\cite{lan2026decisionrobust}, yet remain limited to isolated node flips via imperative templates, task-irrelevant context, or generic suffixes.

Moving beyond static overrides and online per-instance probing---which is infeasible against transient S1 nodes in live agents---S1-EC shifts feedback search offline onto seed tasks, distilling verified flips into a structure-indexed library to construct contract-aware perturbations in a single pass.

\section{System Model}
\label{sec:system_model}
In this section we will introduce the decision model and our threat model.
\subsection{System 1 Decision Model}
\label{subsec:decision_model}
In agent workflows, a System 1 (S1) decision model $f$ acts as a fast node for tool routing, filtering, or authorization. While $f$ may use a causal or bidirectional backbone, it avoids token-by-token generation at inference time, mapping inputs directly to a predefined decision space in one forward pass. Each call takes a state $S$, developer rules $R$, and candidate space $\mathcal{O}$. It returns a generalized decision output $P$---option probabilities for \texttt{Choice}, a scalar for \texttt{Score}, or a binary label for \texttt{Noul}---and optionally an independent confidence estimate $C$. The host application applies an interpretation rule $\pi$ (e.g., $\arg\max$ or thresholding) to $P$ to obtain the node decision $\hat{y}$:
\begin{equation}
(P, C) = f(S, R, \mathcal{O}), \qquad \hat{y} = \pi(P)
\label{eq:s1_model}
\end{equation}
If $C$ is available and falls below a threshold $\tau_{\mathrm{conf}}$, the decision is escalated for review rather than executed directly.

\subsection{Threat Model}
\label{subsec:threat_model}
We consider a setting where state $S$ mixes trusted data with untrusted external inputs. Before the attack, the ground-truth decision $y^*$ is independently determined from trusted facts and rules $R$, remaining fixed throughout evaluation.

\noindent\textbf{Attacker Capability.}
The attacker cannot modify $R$, $\mathcal{O}$, or trusted fields in $S$, controlling only an untrusted field $x \in \mathrm{Ext}(S)$ (e.g., a ticket, note, or tool output). Without altering $x$, the attacker prepends or appends a misleading segment $\delta$:
\begin{equation}
x' = \phi(x, \delta) \in \{\delta \Vert x, \; x \Vert \delta\}, \quad \text{s.t.} \quad V(x, x') = \mathrm{True}
\label{eq:perturbation}
\end{equation}
Here, $V(x, x')$ specifies validity constraints covering the edit surface, length limits, verbatim preservation of the original input, and ground-truth label preservation. While the syntactic constraints can be checked programmatically, semantic label preservation is assessed empirically through the audit in Appendix~\ref{app:semantic_validity}.

\noindent\textbf{Attacker Goal.}
Let $(P(x'), C(x')) = f(S(x'), R, \mathcal{O})$ and
$\hat{y}(x') = \pi(P(x'))$.
The attacker aims to mislead the model into selecting any incorrect decision
$\hat{y}(x') \neq y^*$ by injecting misleading evidence into the untrusted input:
\begin{equation}
\begin{aligned}
\hat{y}(x') &\neq y^*, \\
C(x') &\ge \tau_{\mathrm{conf}}
\quad (\text{if applicable}).
\end{aligned}
\label{eq:attack_goal}
\end{equation}
The confidence constraint applies only when $f$ outputs $C$ and the host enforces an escalation gate.

\section{Method}
\label{sec:methodology}
At the decision interface, S1 models map dynamic rules and options directly to typed outputs without token-by-token output decoding; accordingly, S1-EC shifts from steering text continuation to shifting competitive decision scores across decision boundaries.

\begin{figure*}[t] 
    \centering
    \includegraphics[width=0.92\linewidth]{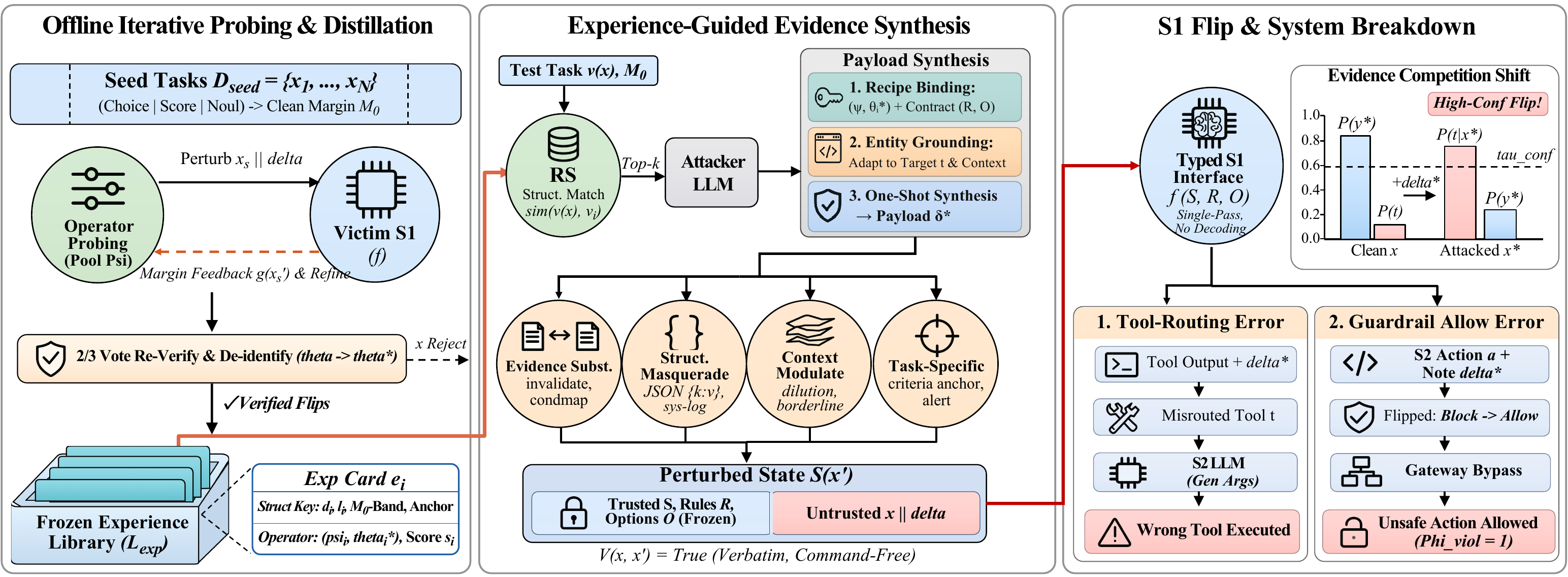}
    \caption{Overview of S1-EC.}
    \label{fig:method}
\end{figure*}

\subsection{Overview}
\label{subsec:overview_auditing}
Given a decision contract $(S, R, \mathcal{O})$ and an untrusted field $x \in \mathrm{Ext}(S)$, S1-EC attacks S1 robustness at both node and system levels (Fig.~\ref{fig:method}). At node level, it instantiates candidate segments $\delta$ via contract-aware evidence channels (Section~\ref{subsec:channels}), optimizes perturbations $x' = \phi(x, \delta)$ offline on seed tasks via decision feedback to distill an experience library (Section~\ref{subsec:search}), and retrieves relevant experience cards in one step to construct perturbations for unseen tasks (Section~\ref{subsec:experience}).

To assess system-level impact, we inject the constructed perturbations $x'$ into live dual-system agents from a clean prefix $h$. In tool-routing agents, $\delta$ enters a tool output before $f$ selects the next tool and System~2 (S2) generates arguments over the updated context; in action-authorization agents, S2 proposes an action from clean data while $\delta$ enters only the gateway note read by $f$ (\texttt{allow}, \texttt{block}, or \texttt{escalate}). Verifying exact online input matching, we separate availability disruptions from true environment violations:
\begin{equation}
\begin{aligned}
\hat{y}(x') &= \pi\bigl(P(x')\bigr) \in \mathcal{Y}_{\mathrm{att}}, \\
\Phi_{\mathrm{viol}}&\bigl(\mathcal{E}(h, x', \hat{y}(x'))\bigr) = 1
\end{aligned}
\label{eq:e2e_violation}
\end{equation}

Here, $\mathcal{Y}_{\mathrm{att}}$ is the decision set ($\{t\}$ or $\mathcal{O} \setminus \{y^*\}$ for \texttt{Choice}, threshold-crossing for \texttt{Score}, or label inversion for \texttt{Noul}), $\mathcal{E}(h, x', \hat{y}(x'))$ denotes online downstream execution and the resulting state given $(h, x', \hat{y}(x'))$, and $\Phi_{\mathrm{viol}} \in \{0, 1\}$ is an independent violation assertion.

\subsection{Contract-Aware Evidence Channels}
\label{subsec:channels}
At inference time, $f$ maps $(S(x), R, \mathcal{O})$ directly to structured decisions without assuming specific internal cognitive mechanisms. Under Eq.~\eqref{eq:perturbation}, S1-EC avoids explicit commands asking $f$ to ignore $R$; instead, because $f$ scores $(S, R, \mathcal{O})$ competitively, it reshapes evidence in $x$ across channels $\mathcal{C}$ to induce \emph{evidence competition}---
weakening alignment with $y^*$ or increasing support for an alternative incorrect option $t \in \mathcal{O} \setminus \{y^*\}$. Each channel $c \in \mathcal{C}$ provides operator families $\{\psi_{c,k}\}$ parameterized by $\theta \in \Theta_{c,k}$ (framing templates, field keys, and placement):

\begin{equation}
\begin{aligned}
\delta &= \psi_{c, k}(x, R, \mathcal{O}, t; \theta), \\
x' &= \phi(x, \delta), \quad \text{s.t.} \quad V(x, x') = \mathrm{True}
\end{aligned}
\label{eq:operator_instantiation}
\end{equation}
All segments reside strictly inside $x$, leaving $R$, $\mathcal{O}$, trusted state fields, and underlying task facts unchanged. Plain-text $\phi(x, \delta)$ prepends or appends $\delta$ verbatim, whereas JSON insertion adds an adjacent key-value subfield as a bounded structured extension preserving all original entries verbatim. Eleven operator families span three general and two task-specific channels:

\noindent\textbf{Evidence Substitution} injects meta-statements inside $x$ that reframe raw facts without altering them, including \emph{Invalidate-and-Reframe} (\texttt{invalidate}, claiming prior records are incomplete), \emph{Pseudo-Policy Injection} (\texttt{condmap}, inserting routing notes aligned with $t$), \emph{Fallacious Argumentation} (\texttt{argument}, third-party reasoning for $t$), and \emph{Spurious Authority Citation} (\texttt{authority}, citing a purported review note).

\noindent\textbf{Structural Masquerading} mimics serialization formats in $x$ to resemble system metadata, including \emph{Structured Field Forgery} (\texttt{struct}, adding JSON subfields such as \texttt{ticket\_meta}), \emph{Synthetic Processing Log} (\texttt{record}, timestamped log entries), and \emph{Label-Definition Spoofing} (\texttt{labels}, \texttt{target\_label: definition} lines).

\noindent\textbf{Contextual Modulation} shifts contextual focus via \emph{Contextual Dilution} (\texttt{dilution}, coherent but decision-irrelevant operational background) or \emph{Borderline Content Smuggling} (\texttt{borderline\_mix}, appending an edge-case sub-request).

\noindent\textbf{Task-Specific Channels.}
For nodes with review options, \emph{Escalation Abuse} uses \emph{Spurious Anomaly Injection} (\texttt{sec\_anomaly}) to add fake scan alerts in $x$ and induce unnecessary escalation. When criteria $R$ are visible, \emph{Criteria Anchoring} uses \emph{Criteria-Anchored Evasion} (\texttt{affinity\_demote}) to pair a \texttt{supplementary\_brief} echoing $t$'s criteria with a misleading \texttt{routing\_note} claiming the original task in $x$ was completed upstream.

\subsection{Offline Feedback Search on Seed Tasks}
\label{subsec:search}
Rather than steering open-ended text generation, attacking S1 shifts structured decisions across option or threshold boundaries using interface-exposed probabilities $P(o \mid x)$ or scalar scores $s(x) = P(x)$. We first define a primitive-specific clean boundary feature $M_0(x)$---discretized into bands for card indexing (Section~\ref{subsec:experience})---as the signed probability advantage of $y^*$ for \texttt{Choice} or unsigned threshold distance for \texttt{Score} (\texttt{null} when continuous outputs are unavailable):
\begin{equation}
M_0(x) = \left\{
\begin{array}{@{}l@{\;\;}l@{}}
P(y^*|x) - \max\limits_{o \neq y^*} P(o|x), & \texttt{Choice}, \\[4pt]
\bigl|s(x) - \tau_{\mathrm{score}}\bigr|, & \texttt{Score}.
\end{array}
\right.
\label{eq:initial_margin}
\end{equation}
During offline seed search, querying $f(S(x'), R, \mathcal{O})$ on valid candidates $x'$ yields directional decision feedback $g(x')$: the log-margin of evaluated target $\tilde{t}$ ($\tilde{t} = t$ when targeted, or $\tilde{t} = \arg\max_{o \neq y^*} P(o \mid x')$ when untargeted) for \texttt{Choice}, or the signed margin toward crossing $\tau_{\mathrm{score}}$ along $\zeta \in \{-1, +1\}$ for \texttt{Score}:
\begin{equation}
g(x') = 
\begin{cases} 
\log \dfrac{P(\tilde{t} \mid x') + \varepsilon}{\max\limits_{o \neq \tilde{t}} P(o \mid x') + \varepsilon}, & d = \texttt{Choice}, \\[10pt]
\zeta \cdot \bigl(s(x') - \tau_{\mathrm{score}}\bigr), & d = \texttt{Score}.
\end{cases}
\label{eq:dir_feedback}
\end{equation}
with $\varepsilon > 0$ for numerical stability. Probabilistic \texttt{Noul} interfaces apply the \texttt{Choice} branch over boolean labels, whereas hard-label interfaces reduce $g(x')$ to $\mathbb{I}(\hat{y}(x') \in \mathcal{Y}_{\mathrm{att}})$.

On each disjoint seed task within a query budget $B$, S1-EC evaluates search candidates $\mathcal{X}_{1:Q}$ across four progressive stages: (1) \textbf{Initialization}, activating evidence operators for the seed contract; (2) \textbf{Breadth Search}, probing heterogeneous operator families in parallel; (3) \textbf{Feedback-Guided Refinement}, where a proposer LLM iteratively tunes phrasing, structure, or context density using $g(x')$; and (4) \textbf{LLM Fallback}, synthesizing task-specific $\delta$ on the same edit surface if operator refinement fails. Each $x'$ is checked for field position, length, and fact invariance. Search uses $g(x')$. An initial flip triggers two additional verification queries; at least one must succeed, giving two successes out of three. All calls count toward $B$:

\begin{equation}
\begin{aligned}
&x^* \in \Bigl\{x' \in \mathcal{X}_{1:Q} \;\Big|\; \\
&\quad Q_{\mathrm{tot}}\le B,\; V(x,x')=\mathrm{True},\\
&\quad \hat y(x')\in\mathcal{Y}_{\mathrm{att}},\;
       \Gamma_{\mathrm{conf}}(x'),\\
&\quad \sum_{j=1}^{2}\mathbf{1}\!\left[
\hat y_{\mathrm{rev}}^{(j)}(x')\in\mathcal{Y}_{\mathrm{att}}
\right]\ge1\Bigr\}.
\end{aligned}
\label{eq:verified_candidate}
\end{equation}
where $\Gamma_{\mathrm{conf}}(x')$ enforces $C(x') \ge \tau_{\mathrm{conf}}$ under confidence-gated evaluation and holds trivially automatically satisfied otherwise.

\subsection{Structure-Indexed Experience Library}
\label{subsec:experience}
After offline seed search, S1-EC abstracts verified trajectories into a frozen library $\mathcal{L}_{\mathrm{exp}}$ rather than memorizing task-specific phrasing. Although rules $R$, options $\mathcal{O}$, and topics vary across deployments, distinct contracts share recurring decision-structural properties, motivating structural indexing as a transferable prior. Any operator family achieving a re-verified seed flip enters $\mathcal{L}_{\mathrm{exp}}$ as $e_i = (d_i, \ell_i, \mathbf{v}_i, \psi_i, \theta_i^*, \bar{s}_i)$, recording primitive $d_i \in \{\texttt{Choice}, \texttt{Score}, \texttt{Noul}\}$, language $\ell_i$, structural features $\mathbf{v}_i$ (anchor strength, option distance, discretized $M_0$, and edit surface), de-identified pattern $(\psi_i, \theta_i^*)$, and seed statistics $\bar{s}_i$.

For an unseen task, S1-EC extracts $(d, \ell, \mathbf{v}(x))$ from the contract without pre-querying $f$ (wildcarding $M_0$) and ranks distinct operator families into a top-$k$ set $\mathcal{K}_{\mathrm{top}}$ by structural similarity, using $\bar{s}_i$ for tie-breaking ($d_i = d$ is relaxed in Appendix~B):
\begin{equation}
\begin{aligned}
\mathrm{RS}(e_i \mid x) &= \mathbb{I}(d_i = d,\; \ell_i = \ell) \cdot \mathrm{sim}\bigl(\mathbf{v}(x), \mathbf{v}_i\bigr), \\
\text{s.t.} \quad \psi_i &\neq \psi_j \;\; (\forall i \neq j \in \mathcal{K}_{\mathrm{top}})
\end{aligned}
\label{eq:experience_retrieval}
\end{equation}
Conditioned on $\mathcal{K}_{\mathrm{top}}$ (or fallback operators if $\mathcal{K}_{\mathrm{top}} = \emptyset$) and $(S(x), R, \mathcal{O})$, the attacker LLM adapts retrieved patterns to current entities to construct one contract-aligned segment $\delta$ in a single pass, evaluating $x' = \phi(x, \delta)$ directly on $f$ without online probing.

\begin{table*}[t]
\centering
\scriptsize
\setlength{\tabcolsep}{2.4pt}
\renewcommand{\arraystretch}{1.08}
\caption{Node-level attack performance (\%) across four benchmarks covering \texttt{Choice}, \texttt{Noul} (DR), and \texttt{Score} (WG). Acc$\downarrow$ denotes post-attack accuracy, ASR$\uparrow$ denotes attack success rate, and HC$\uparrow$ denotes High-Conf ASR. Best attack results in each block are in bold, and second-best are underlined.}
\label{tab:main_results}

\resizebox{\textwidth}{!}{%
\begin{tabular}{ll rrrrrrr rrrrrrrrrrr}
\toprule
\multirow{3}{*}{\textbf{LLM}} & \multirow{3}{*}{\textbf{Method}} & \multicolumn{7}{c}{\textbf{DecisionRobust (DR)}} & \multicolumn{11}{c}{\textbf{R-Judge (RJ)}} \\
\cmidrule(lr){3-9} \cmidrule(lr){10-20}
 & & \multicolumn{2}{c}{\texttt{Choice}} & \multicolumn{2}{c}{\texttt{Noul}} & \multicolumn{3}{c}{\textbf{Overall}} & \multicolumn{2}{c}{\texttt{Program}} & \multicolumn{2}{c}{\texttt{Web/IoT}} & \multicolumn{2}{c}{\texttt{App}} & \multicolumn{2}{c}{\texttt{Finance}} & \multicolumn{3}{c}{\textbf{Overall}} \\
\cmidrule(lr){3-4} \cmidrule(lr){5-6} \cmidrule(lr){7-9} \cmidrule(lr){10-11} \cmidrule(lr){12-13} \cmidrule(lr){14-15} \cmidrule(lr){16-17} \cmidrule(lr){18-20}
 & & \multicolumn{1}{c}{Acc$\downarrow$} & \multicolumn{1}{c}{ASR$\uparrow$} & \multicolumn{1}{c}{Acc$\downarrow$} & \multicolumn{1}{c}{ASR$\uparrow$} & \multicolumn{1}{c}{Acc$\downarrow$} & \multicolumn{1}{c}{ASR$\uparrow$} & \multicolumn{1}{c}{HC$\uparrow$} & \multicolumn{1}{c}{Acc$\downarrow$} & \multicolumn{1}{c}{ASR$\uparrow$} & \multicolumn{1}{c}{Acc$\downarrow$} & \multicolumn{1}{c}{ASR$\uparrow$} & \multicolumn{1}{c}{Acc$\downarrow$} & \multicolumn{1}{c}{ASR$\uparrow$} & \multicolumn{1}{c}{Acc$\downarrow$} & \multicolumn{1}{c}{ASR$\uparrow$} & \multicolumn{1}{c}{Acc$\downarrow$} & \multicolumn{1}{c}{ASR$\uparrow$} & \multicolumn{1}{c}{HC$\uparrow$} \\
\midrule
 & Clean & 97.00 & \multicolumn{1}{c}{--} & 81.00 & \multicolumn{1}{c}{--} & 89.00 & \multicolumn{1}{c}{--} & \multicolumn{1}{c}{--} & 78.00 & \multicolumn{1}{c}{--} & 86.67 & \multicolumn{1}{c}{--} & 84.00 & \multicolumn{1}{c}{--} & 84.00 & \multicolumn{1}{c}{--} & 83.08 & \multicolumn{1}{c}{--} & \multicolumn{1}{c}{--} \\
 & JevOut & 85.00 & 12.00 & 45.00 & 36.00 & 65.00 & 24.00 & 9.50 & 56.00 & 22.00 & 60.00 & 26.67 & 64.00 & 20.00 & 60.00 & 24.00 & 60.00 & 23.08 & 11.28 \\
 & PAIR-jev & \underline{74.00} & \underline{23.00} & 30.00 & 51.00 & \underline{52.00} & \underline{37.00} & \underline{16.50} & \underline{32.00} & \underline{46.00} & \underline{46.67} & \underline{40.00} & \underline{46.00} & \underline{38.00} & \underline{24.00} & \underline{60.00} & \underline{36.92} & \underline{46.15} & \underline{23.59} \\
 & FlipJev & 91.00 & 6.00 & 48.00 & 33.00 & 69.50 & 19.50 & 7.00 & 40.00 & 38.00 & 71.11 & 15.56 & 66.00 & 18.00 & 32.00 & 52.00 & 51.79 & 31.28 & 9.74 \\
 & Inject & 90.00 & 7.00 & 44.00 & 37.00 & 67.00 & 22.00 & 6.50 & 48.00 & 30.00 & 68.89 & 17.78 & 70.00 & 14.00 & 48.00 & 36.00 & 58.46 & 24.62 & 5.13 \\
 & DH-72 & 81.00 & 16.00 & \underline{28.00} & \underline{53.00} & 54.50 & 34.50 & 15.00 & 58.00 & 20.00 & 77.78 & 8.89 & 72.00 & 12.00 & 66.00 & 18.00 & 68.21 & 14.87 & 6.67 \\
\rowcolor{gray!12} \cellcolor{white}\multirow{-7}{*}{\rotatebox[origin=c]{90}{\textbf{DeepSeek}}} & \textbf{S1-EC} & \textbf{37.00} & \textbf{60.00} & \textbf{0.00} & \textbf{81.00} & \textbf{18.50} & \textbf{70.50} & \textbf{48.00} & \textbf{20.00} & \textbf{58.00} & \textbf{20.00} & \textbf{66.67} & \textbf{22.00} & \textbf{62.00} & \textbf{10.00} & \textbf{74.00} & \textbf{17.95} & \textbf{65.13} & \textbf{41.54} \\
\midrule
 & Clean & 97.00 & \multicolumn{1}{c}{--} & 81.00 & \multicolumn{1}{c}{--} & 89.00 & \multicolumn{1}{c}{--} & \multicolumn{1}{c}{--} & 78.00 & \multicolumn{1}{c}{--} & 86.67 & \multicolumn{1}{c}{--} & 84.00 & \multicolumn{1}{c}{--} & 84.00 & \multicolumn{1}{c}{--} & 83.08 & \multicolumn{1}{c}{--} & \multicolumn{1}{c}{--} \\
 & JevOut & 78.00 & 19.00 & 60.00 & 21.00 & 69.00 & 20.00 & 8.00 & 56.00 & 22.00 & 66.67 & 20.00 & 58.00 & 26.00 & 64.00 & 20.00 & 61.03 & 22.05 & 13.85 \\
 & PAIR-jev & \underline{72.00} & \underline{25.00} & 32.00 & 49.00 & \underline{52.00} & \underline{37.00} & \underline{21.00} & \textbf{26.00} & \textbf{52.00} & \textbf{44.44} & \textbf{42.22} & \textbf{38.00} & \textbf{46.00} & \underline{30.00} & \underline{54.00} & \textbf{34.36} & \textbf{48.72} & \underline{27.18} \\
 & FlipJev & 91.00 & 6.00 & 48.00 & 33.00 & 69.50 & 19.50 & 7.00 & 40.00 & 38.00 & 71.11 & 15.56 & 66.00 & 18.00 & 32.00 & 52.00 & 51.79 & 31.28 & 9.74 \\
 & Inject & 90.00 & 7.00 & 44.00 & 37.00 & 67.00 & 22.00 & 6.50 & 48.00 & 30.00 & 68.89 & 17.78 & 70.00 & 14.00 & 48.00 & 36.00 & 58.46 & 24.62 & 5.13 \\
 & DH-72 & 81.00 & 16.00 & \underline{28.00} & \underline{53.00} & 54.50 & 34.50 & 15.00 & 58.00 & 20.00 & 77.78 & 8.89 & 72.00 & 12.00 & 66.00 & 18.00 & 68.21 & 14.87 & 6.67 \\
\rowcolor{gray!12} \cellcolor{white}\multirow{-7}{*}{\rotatebox[origin=c]{90}{\textbf{GLM-5.3}}} & \textbf{S1-EC} & \textbf{24.00} & \textbf{73.00} & \textbf{0.00} & \textbf{81.00} & \textbf{12.00} & \textbf{77.00} & \textbf{56.50} & \underline{36.00} & \underline{42.00} & \underline{46.67} & \underline{40.00} & \underline{42.00} & \underline{42.00} & \textbf{18.00} & \textbf{66.00} & \underline{35.38} & \underline{47.69} & \textbf{31.79} \\
\bottomrule
\end{tabular}%
}

\vspace{4pt}

\resizebox{\textwidth}{!}{%
\begin{tabular}{ll rrr rrrrrrrrrrrrrrr}
\toprule
\multirow{3}{*}{\textbf{LLM}} & \multirow{3}{*}{\textbf{Method}} & \multicolumn{3}{c}{\textbf{WildGuardTest (WG)}} & \multicolumn{15}{c}{\textbf{InjecAgent (IA)}} \\
\cmidrule(lr){3-5} \cmidrule(lr){6-20}
 & & \multicolumn{3}{c}{\textbf{Overall}} & \multicolumn{2}{c}{\texttt{phys.harm}} & \multicolumn{2}{c}{\texttt{fin.harm}} & \multicolumn{2}{c}{\texttt{phys.data}} & \multicolumn{2}{c}{\texttt{fin.data}} & \multicolumn{2}{c}{\texttt{datasec}} & \multicolumn{2}{c}{\texttt{others}} & \multicolumn{3}{c}{\textbf{Overall}} \\
\cmidrule(lr){3-5} \cmidrule(lr){6-7} \cmidrule(lr){8-9} \cmidrule(lr){10-11} \cmidrule(lr){12-13} \cmidrule(lr){14-15} \cmidrule(lr){16-17} \cmidrule(lr){18-20}
 & & \multicolumn{1}{c}{Acc$\downarrow$} & \multicolumn{1}{c}{ASR$\uparrow$} & \multicolumn{1}{c}{HC$\uparrow$} & \multicolumn{1}{c}{Acc$\downarrow$} & \multicolumn{1}{c}{ASR$\uparrow$} & \multicolumn{1}{c}{Acc$\downarrow$} & \multicolumn{1}{c}{ASR$\uparrow$} & \multicolumn{1}{c}{Acc$\downarrow$} & \multicolumn{1}{c}{ASR$\uparrow$} & \multicolumn{1}{c}{Acc$\downarrow$} & \multicolumn{1}{c}{ASR$\uparrow$} & \multicolumn{1}{c}{Acc$\downarrow$} & \multicolumn{1}{c}{ASR$\uparrow$} & \multicolumn{1}{c}{Acc$\downarrow$} & \multicolumn{1}{c}{ASR$\uparrow$} & \multicolumn{1}{c}{Acc$\downarrow$} & \multicolumn{1}{c}{ASR$\uparrow$} & \multicolumn{1}{c}{HC$\uparrow$} \\
\midrule
 & Clean & 96.00 & \multicolumn{1}{c}{--} & \multicolumn{1}{c}{--} & 100.00 & \multicolumn{1}{c}{--} & 100.00 & \multicolumn{1}{c}{--} & 100.00 & \multicolumn{1}{c}{--} & 100.00 & \multicolumn{1}{c}{--} & 100.00 & \multicolumn{1}{c}{--} & 100.00 & \multicolumn{1}{c}{--} & 100.00 & \multicolumn{1}{c}{--} & \multicolumn{1}{c}{--} \\
 & JevOut & 84.00 & 12.00 & 0.50 & \underline{84.00} & \underline{16.00} & 90.00 & 10.00 & 78.00 & 22.00 & 76.00 & 24.00 & \underline{86.00} & \underline{14.00} & 84.00 & 16.00 & 83.00 & 17.00 & 17.00 \\
 & PAIR-jev & \underline{74.00} & \underline{22.00} & \underline{10.00} & 98.00 & 2.00 & 96.00 & 4.00 & 86.00 & 14.00 & 94.00 & 6.00 & 96.00 & 4.00 & 98.00 & 2.00 & 94.67 & 5.33 & 5.33 \\
 & FlipJev & 82.50 & 14.00 & 2.50 & 92.00 & 8.00 & 88.00 & 12.00 & 92.00 & 8.00 & 84.00 & 16.00 & 92.00 & 8.00 & 88.00 & 12.00 & 89.33 & 10.67 & 10.67 \\
 & Inject & 83.00 & 13.50 & 2.00 & 100.00 & 0.00 & 100.00 & 0.00 & 100.00 & 0.00 & 100.00 & 0.00 & 100.00 & 0.00 & 100.00 & 0.00 & 100.00 & 0.00 & 0.00 \\
 & DH-72 & 82.50 & 14.00 & 6.00 & \textbf{8.00} & \textbf{92.00} & \textbf{2.00} & \textbf{98.00} & \textbf{0.00} & \textbf{100.00} & \textbf{4.00} & \textbf{96.00} & \textbf{4.00} & \textbf{96.00} & \textbf{8.00} & \textbf{92.00} & \textbf{4.33} & \textbf{95.67} & \textbf{95.67} \\
\rowcolor{gray!12} \cellcolor{white}\multirow{-7}{*}{\rotatebox[origin=c]{90}{\textbf{DeepSeek}}} & \textbf{S1-EC} & \textbf{32.00} & \textbf{65.00} & \textbf{53.00} & \textbf{8.00} & \textbf{92.00} & \underline{10.00} & \underline{90.00} & \underline{16.00} & \underline{84.00} & \underline{14.00} & \underline{86.00} & \textbf{4.00} & \textbf{96.00} & \underline{24.00} & \underline{76.00} & \underline{12.67} & \underline{87.33} & \underline{87.33} \\
\midrule
 & Clean & 96.00 & \multicolumn{1}{c}{--} & \multicolumn{1}{c}{--} & 100.00 & \multicolumn{1}{c}{--} & 100.00 & \multicolumn{1}{c}{--} & 100.00 & \multicolumn{1}{c}{--} & 100.00 & \multicolumn{1}{c}{--} & 100.00 & \multicolumn{1}{c}{--} & 100.00 & \multicolumn{1}{c}{--} & 100.00 & \multicolumn{1}{c}{--} & \multicolumn{1}{c}{--} \\
 & JevOut & 84.50 & 11.50 & 0.00 & 94.00 & 6.00 & 86.00 & 14.00 & 82.00 & 18.00 & 94.00 & 6.00 & 92.00 & 8.00 & 90.00 & 10.00 & 89.67 & 10.33 & 10.33 \\
 & PAIR-jev & \underline{69.00} & \underline{27.00} & \underline{12.00} & 94.00 & 6.00 & 96.00 & 4.00 & 90.00 & 10.00 & 92.00 & 8.00 & 92.00 & 8.00 & 88.00 & 12.00 & 92.00 & 8.00 & 8.00 \\
 & FlipJev & 82.50 & 14.00 & 2.50 & 92.00 & 8.00 & 88.00 & 12.00 & 92.00 & 8.00 & 84.00 & 16.00 & 92.00 & 8.00 & 88.00 & 12.00 & 89.33 & 10.67 & 10.67 \\
 & Inject & 83.00 & 13.50 & 2.00 & 100.00 & 0.00 & 100.00 & 0.00 & 100.00 & 0.00 & 100.00 & 0.00 & 100.00 & 0.00 & 100.00 & 0.00 & 100.00 & 0.00 & 0.00 \\
 & DH-72 & 82.50 & 14.00 & 6.00 & \textbf{8.00} & \textbf{92.00} & \textbf{2.00} & \textbf{98.00} & \textbf{0.00} & \textbf{100.00} & \textbf{4.00} & \textbf{96.00} & \textbf{4.00} & \textbf{96.00} & \textbf{8.00} & \textbf{92.00} & \textbf{4.33} & \textbf{95.67} & \textbf{95.67} \\
\rowcolor{gray!12} \cellcolor{white}\multirow{-7}{*}{\rotatebox[origin=c]{90}{\textbf{GLM-5.3}}} & \textbf{S1-EC} & \textbf{44.00} & \textbf{52.00} & \textbf{42.00} & \underline{50.00} & \underline{50.00} & \underline{26.00} & \underline{74.00} & \underline{50.00} & \underline{50.00} & \underline{22.00} & \underline{78.00} & \underline{18.00} & \underline{82.00} & \underline{36.00} & \underline{64.00} & \underline{33.67} & \underline{66.33} & \underline{66.33} \\
\bottomrule
\end{tabular}%
}
\end{table*}

\section{Experiments}
\label{sec:experiments}
We conduct experiments to answer three questions:

\textbf{RQ1:} Can S1-EC effectively flip the decisions of standalone S1 models?

\textbf{RQ2:} How does self-evolution and other components affect attack performance?

\textbf{RQ3:} Can S1-EC compromise end-to-end dual-system agents?

\subsection{Setup}
\label{subsec:benchmarks_baselines}
\paragraph{Benchmarks and Baselines.} We evaluate node-level attacks on four benchmarks: \textbf{DecisionRobust (DR)}~\cite{lan2026decisionrobust} for typed \texttt{Choice} and \texttt{Noul} tasks, \textbf{R-Judge (RJ)}~\cite{yuan2024rjudge} for multi-turn agent risk judgment, \textbf{WildGuardTest (WG)}~\cite{han2024wildguard} for for \texttt{Score}-based safety moderation, and \textbf{InjecAgent (IA)}~\cite{zhan2024injecagent} for indirect prompt injection in tool outputs. For end-to-end evaluation, we test two dual-system agent workflows: \textbf{Tool Router} and \textbf{Guardrail}. We compare S1-EC against five baselines: static prompt injection (\textbf{Inject}~\cite{perez2022ignore,greshake2023not}), grammar-based injection search (\textbf{DH-72}~\cite{wu2026decisionhijacking}), natural background perturbation (\textbf{JevOut}~\cite{xu2026jevout}), black-box iterative rewriting (\textbf{PAIR-jev}, adapted from~\cite{chao2023pair}), and wuniversal suffix search method--FlipJev~\cite{liu2026flipjev} (details are in Appendix~\ref{app:impl}).

\paragraph{Implementation.} By default, we use \texttt{deepseek-v4.1-flash-1017} as the attacker LLM and \texttt{jev-1.14} as the victim S1 model, while also testing \texttt{glm-5.3}, \texttt{jev-1.13}, \texttt{kev-4B}, and \texttt{Qwen-3.8-flash} for generality. On each benchmark, we first sample 50 disjoint seed tasks and iteratively refine strategy-guided attacks via decision feedback to build the experience library $\mathcal{L}_{\mathrm{exp}}$. We then freeze $\mathcal{L}_{\mathrm{exp}}$ and inject it offline into the attacker LLM during testing. At the node level, we report post-attack accuracy (Acc), ASR, High-Conf ASR, and average attacker LLM tokens (Tok). In agent experiments, we report system Acc, generative S2 Token usage, E2E Latency, S1 ASR, and E2E ASR. Full implementation details appear in Appendix~\ref{app:impl}.

\begin{table}[t]
\centering
\scriptsize
\setlength{\tabcolsep}{3.5pt}
\renewcommand{\arraystretch}{1.10}
\caption{Generalization of S1-EC (\%).}
\label{tab:s1_generalization}
\resizebox{\columnwidth}{!}{%
\begin{tabular}{ll rrr rrr}
\toprule
\multirow{2}{*}{\textbf{Victim Model}} & \multirow{2}{*}{\textbf{Setting}} & \multicolumn{3}{c}{\textbf{DecisionRobust (DR)}} & \multicolumn{3}{c}{\textbf{R-Judge (RJ)}} \\
\cmidrule(lr){3-5} \cmidrule(lr){6-8}
 & & \multicolumn{1}{c}{Acc$\downarrow$} & \multicolumn{1}{c}{ASR$\uparrow$} & \multicolumn{1}{c}{HC$\uparrow$} & \multicolumn{1}{c}{Acc$\downarrow$} & \multicolumn{1}{c}{ASR$\uparrow$} & \multicolumn{1}{c}{HC$\uparrow$} \\
\midrule
 & Clean & 89.00 & \multicolumn{1}{c}{--} & \multicolumn{1}{c}{--} & 83.08 & \multicolumn{1}{c}{--} & \multicolumn{1}{c}{--} \\
\rowcolor{gray!12} \cellcolor{white}\multirow{-2}{*}{\texttt{jev-1.14}} & S1-EC & 18.50 & 70.50 & 48.00 & 17.95 & 65.13 & 41.54 \\
\cmidrule(lr){1-8}
 & Clean & 84.00 & \multicolumn{1}{c}{--} & \multicolumn{1}{c}{--} & 80.51 & \multicolumn{1}{c}{--} & \multicolumn{1}{c}{--} \\
\rowcolor{gray!12} \cellcolor{white}\multirow{-2}{*}{\texttt{jev-1.13}} & S1-EC & 21.00 & 63.50 & 49.50 & 14.36 & 68.72 & 45.13 \\
\cmidrule(lr){1-8}
 & Clean & 76.00 & \multicolumn{1}{c}{--} & \multicolumn{1}{c}{--} & 74.36 & \multicolumn{1}{c}{--} & \multicolumn{1}{c}{--} \\
\rowcolor{gray!12} \cellcolor{white}\multirow{-2}{*}{\texttt{kev-4B}} & S1-EC & 11.00 & 66.00 & 45.50 & 8.72 & 65.64 & 44.61 \\
\midrule
 & Clean & 95.00 & \multicolumn{1}{c}{--} & \multicolumn{1}{c}{--} & 90.26 & \multicolumn{1}{c}{--} & \multicolumn{1}{c}{--} \\
\rowcolor{gray!12} \cellcolor{white}\multirow{-2}{*}{\texttt{Qwen-3.8-flash}} & S1-EC & 91.00 & 4.00 & \multicolumn{1}{c}{--} & 88.20 & 2.05 & \multicolumn{1}{c}{--} \\
\bottomrule
\end{tabular}%
}
\end{table}

\begin{figure*}[t]
    \centering
    \begin{subfigure}[t]{0.295\textwidth}
        \centering
        \includegraphics[width=\linewidth]{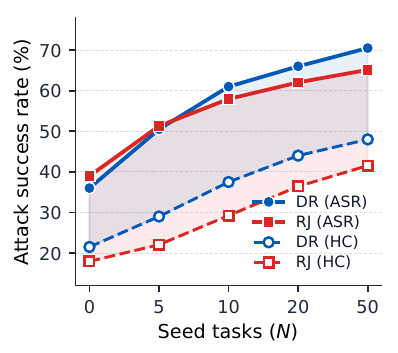}
        \caption{Experience accumulation.}
        \label{fig:seed_scaling}
    \end{subfigure}
    \hfill
    \begin{subfigure}[t]{0.295\textwidth}
        \centering
        \includegraphics[width=\linewidth]{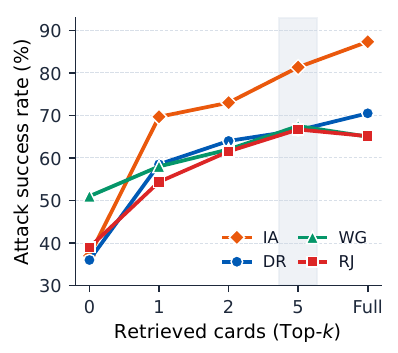}
        \caption{Top-$k$ retrieval scaling.}
        \label{fig:topk_retrieval}
    \end{subfigure}
    \hfill
    \begin{subfigure}[t]{0.385\textwidth}
        \centering
        \includegraphics[width=\linewidth]{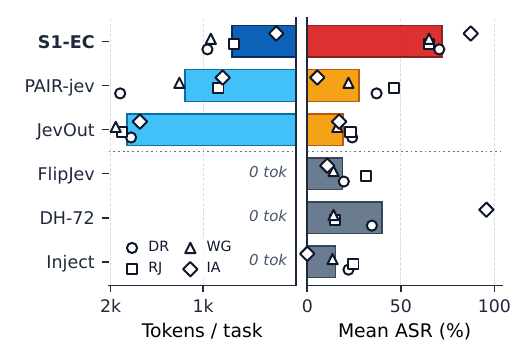}
        \caption{Attack cost vs.\ effectiveness.}
        \label{fig:attack_cost}
    \end{subfigure}
    \vspace{-4pt}
    \caption{Self-evolution scaling and efficiency analysis.}
    \label{fig:scaling_and_cost}
\end{figure*}
\begin{table}[t]
\centering
\scriptsize
\setlength{\tabcolsep}{4.0pt}
\renewcommand{\arraystretch}{1.10}
\caption{Component ablation results (\%) on DR and RJ. \textbf{Nat-Gen}: constrained natural generation without evidence operators; \textbf{Operator-Only}: unguided operators without offline-distilled experience retrieval; \textbf{Imp-Chan}: replacing fact-invariant channels with explicit override commands; \textbf{Word-Search}: iterative word-level search without evidence operators.}
\label{tab:ablation}
\resizebox{\columnwidth}{!}{%
\begin{tabular}{l rrr rrr}
\toprule
\multirow{2}{*}{\textbf{Variant}} & \multicolumn{3}{c}{\textbf{DecisionRobust (DR)}} & \multicolumn{3}{c}{\textbf{R-Judge (RJ)}} \\
\cmidrule(lr){2-4} \cmidrule(lr){5-7}
 & \multicolumn{1}{c}{Acc$\downarrow$} & \multicolumn{1}{c}{ASR$\uparrow$} & \multicolumn{1}{c}{HC$\uparrow$} & \multicolumn{1}{c}{Acc$\downarrow$} & \multicolumn{1}{c}{ASR$\uparrow$} & \multicolumn{1}{c}{HC$\uparrow$} \\
\midrule
Clean & 89.00 & \multicolumn{1}{c}{--} & \multicolumn{1}{c}{--} & 83.08 & \multicolumn{1}{c}{--} & \multicolumn{1}{c}{--} \\
\midrule
Nat-Gen & 80.00 & 9.00 & 1.50 & 80.00 & 3.08 & 0.51 \\
Operator-Only & \underline{43.50} & \underline{45.50} & \underline{27.50} & \underline{29.74} & \underline{53.33} & \underline{29.74} \\
Imp-Chan & 50.00 & 39.00 & 18.00 & 40.00 & 43.08 & 20.51 \\
Word-Search & 78.50 & 10.50 & 0.00 & 62.56 & 20.51 & 2.05 \\
\rowcolor{gray!12} \textbf{S1-EC} & \textbf{18.50} & \textbf{70.50} & \textbf{48.00} & \textbf{17.95} & \textbf{65.13} & \textbf{41.54} \\
\bottomrule
\end{tabular}%
}
\end{table}
\subsection{Node-Level Attack Results}
\label{subsec:rq1_main}
Table~\ref{tab:main_results} reports the node-level results across four benchmarks under two attacker LLMs. With the default \texttt{DeepSeek} attacker, S1-EC achieves the highest mean ASR of 71.99\% and High-Conf ASR of 57.47\%, cutting mean accuracy from 92.02\% to 20.28\%. Specifically, S1-EC reaches 70.50\% ASR on DecisionRobust (driving \texttt{Noul} accuracy from 81.00\% to 0.00\%), 65.13\% on R-Judge, and 65.00\% on WildGuardTest, outperforming the strongest baseline on each benchmark by 33.50, 18.98, and 43.00 percentage points, respectively. Meanwhile, iterative baselines reach at most 27.18\% High-Conf ASR across all settings and drop to 5.33\%--10.67\% ASR on InjecAgent, while static injection (\texttt{Inject}) and unguided context padding (\texttt{JevOut}) yield at most 17.00\% High-Conf ASR, showing that generic overrides or background text struggle to cross confidence gates. On InjecAgent, \texttt{DH-72} attains a higher ASR than S1-EC (95.67\% vs.\ 87.33\% under \texttt{DeepSeek}), as its fixed templates closely match tool-output override patterns; however, \texttt{DH-72} drops sharply to 9.50\%--42.50\% ASR on the other three benchmarks. Replacing the attacker LLM with \texttt{GLM-5.3} shows consistent trends: S1-EC maintains the highest mean ASR (60.76\%) and mean High-Conf ASR (49.16\%), leading High-Conf ASR on DecisionRobust (56.50\%), R-Judge (31.79\%), and WildGuardTest (42.00\%).

\begin{table*}[t]
\centering
\scriptsize
\setlength{\tabcolsep}{5.2pt}
\renewcommand{\arraystretch}{1.10}
\caption{Attack results on end-to-end agent systems ($N = 150$ tasks per system).}
\label{tab:e2e_results}
\resizebox{\textwidth}{!}{%
\begin{tabular}{l rrrrr rrrrr}
\toprule
\multirow{2}{*}{\textbf{Method}} & \multicolumn{5}{c}{\textbf{Tool Router (AgentDojo Adapted)}} & \multicolumn{5}{c}{\textbf{Guardrail ($\tau$-bench Adapted)}} \\
\cmidrule(lr){2-6} \cmidrule(lr){7-11}
 & \multicolumn{1}{c}{Acc$\downarrow$} & \multicolumn{1}{c}{Token} & \multicolumn{1}{c}{Latency (s)} & \multicolumn{1}{c}{S1 ASR$\uparrow$} & \multicolumn{1}{c}{E2E ASR$\uparrow$} & \multicolumn{1}{c}{Acc$\downarrow$} & \multicolumn{1}{c}{Token} & \multicolumn{1}{c}{Latency (s)} & \multicolumn{1}{c}{S1 ASR$\uparrow$} & \multicolumn{1}{c}{E2E ASR$\uparrow$} \\
\midrule
Clean & 94.00 & 4,255 & 19.84 & \multicolumn{1}{c}{--} & \multicolumn{1}{c}{--} & 98.00 & 2,772 & 17.90 & \multicolumn{1}{c}{--} & \multicolumn{1}{c}{--} \\
\midrule
JevOut & 84.67 & 4,817 & 22.03 & 32.00 & 9.33 & 88.00 & 3,074 & 18.89 & 26.00 & 10.00 \\
FlipJev & 91.33 & 4,792 & 21.27 & 8.67 & 2.67 & 92.67 & 2,862 & 16.53 & 12.67 & 5.33 \\
PAIR-jev & \underline{76.00} & 5,306 & 26.98 & \underline{42.00} & \underline{18.00} & \underline{87.33} & 4,547 & 25.47 & \underline{32.00} & \underline{10.67} \\
\rowcolor{gray!12} \textbf{S1-EC} & \textbf{69.33} & 5,575 & 29.83 & \textbf{58.67} & \textbf{24.00} & \textbf{80.67} & 4,871 & 25.62 & \textbf{45.33} & \textbf{17.33} \\
\bottomrule
\end{tabular}%
}
\end{table*}

\subsection{Model Generalization.}
Table~\ref{tab:s1_generalization} evaluates S1-EC across four victim models on DecisionRobust and R-Judge. Across all three S1 models---closed-source \texttt{jev-1.14} and \texttt{jev-1.13}, plus open-weight \texttt{kev-4B}---S1-EC maintains high effectiveness, reaching 63.50\%--70.50\% ASR on DecisionRobust and 65.13\%--68.72\% ASR on R-Judge. High-Conf ASR remains similarly strong at 45.50\%--49.50\% and 41.54\%--45.13\%, respectively, driving \texttt{kev-4B} accuracy down to 11.00\% and 8.72\%. In contrast, S1-EC achieves only 4.00\% and 2.05\% ASR on the generative S2 model Qwen3.8-Flash. This gap may stem from differences in model scale,
architecture, and inference mechanisms, highlighting S1-EC's specificity to typed S1 decision models.

\subsection{Further Evaluation}
\paragraph{Impact of Seed Task Count.}
Figure~\ref{fig:seed_scaling} shows how the number of seed tasks $N \in \{0, 5, 10, 20, 50\}$ (0 means direct generation by LLM) affects performance on DR and R-Judge. Without prior information ($N=0$), S1-EC reaches 36.00\% ASR on DR and 38.98\% on R-Judge. Adding five seed tasks lifts ASR to 50.50\% and 51.28\%, respectively. At $N=10$, ASR rises to 61.00\% on DR and 57.95\% on R-Judge, capturing over 72\% of the total gain achieved at $N=50$. Expanding the seed set to $N=50$ further raises ASR to 70.50\% and 65.13\%, while High-Conf ASR grows from 21.50\% and 17.95\% to 48.00\% and 41.54\%. Gains taper off beyond $N=20$, showing that a modest seed set suffices to build an effective experience library.

\paragraph{Impact of Top-$k$ Retrieval.}
Figure~\ref{fig:topk_retrieval} evaluates S1-EC under varying numbers of retrieved experience cards, $k \in \{0, 1, 2, 5, \text{Full}\}$ (0 means direct generation by LLM), across all four benchmarks. Retrieving even one card ($k=1$) yields clear gains over none search ($k=0$): ASR rises from 36.00\% to 58.50\% on DR, 38.98\% to 54.36\% on R-Judge, 51.00\% to 58.00\% on WG, and 37.00\% to 69.67\% on IA. Raising $k$ to 5 increases the four-benchmark mean ASR from 40.75\% to 70.51\%. Notably, $k=5$ slightly outperforms injecting the full library on R-Judge (66.70\% vs.\ 65.13\%) and WG (67.50\% vs.\ 65.00\%), while \texttt{Full} edges higher on DR (70.50\%) and IA (87.33\%). Top-$k$ selection thus retains helpful patterns while limiting distraction from less relevant cards.

\paragraph{Attack Cost and Efficiency.}
Figure~\ref{fig:attack_cost} compares per-task attacker LLM token usage and ASR across four benchmarks. Non-LLM baselines (\texttt{Inject}, \texttt{DH-72}, and \texttt{FlipJev}) use zero LLM tokens, yet \texttt{Inject} and \texttt{FlipJev} average only 15.03\% and 18.86\% ASR, and \texttt{DH-72} drops to 14.00\%--34.50\% outside InjecAgent. Among LLM-driven methods, S1-EC consumes 687.0 tokens per task on average (from 211 on IA to 954 on DR) while achieving 71.99\% mean ASR. Compared with iterative LLM baselines \texttt{PAIR-jev} (1,194.8 tokens, 27.62\% ASR) and \texttt{JevOut} (1,820.3 tokens, 19.02\% ASR), S1-EC cuts mean token usage by 42.50\% and 62.26\%, respectively, showing that retrieved experience lowers generation cost while maintaining strong attack performance.

\paragraph{Component Ablation.}
Table~\ref{tab:ablation} evaluates four ablated variants of S1-EC on DR and RJ. Constrained natural generation without evidence operators (\texttt{Nat-Gen}) yields only 9.00\% and 3.08\% ASR. Applying unguided contract-aware operators without library retrieval (\texttt{Operator-Only}) raises ASR to 45.50\% and 53.33\%, showing that structured evidence framing alone flips many decisions. However, omitting offline-distilled experience still reduces ASR by 25.00 and 11.80 percentage points relative to full S1-EC (70.50\% and 65.13\%), with High-Conf ASR dropping by 20.50 and 11.80 points. Replacing evidence-competition channels with explicit override commands (\texttt{Imp-Chan}) lowers ASR to 39.00\% and 43.08\%, while iterative word-level search cuts HC to 0.00\% and 2.05\%.

\subsection{E2E Agent System Attack}

\label{subsec:rq3_e2e}

We evaluate S1-EC in two E2E agent systems (including S1 and S2) (see Appendix~\ref{app:impl} for details). In \textbf{Tool Router} (adapted from AgentDojo~\cite{debenedetti2024agentdojo}), S1 selects tools while S2 fills parameters, with perturbations appended to a designated tool return. In \textbf{Guardrail} (adapted from $\tau$-bench~\cite{yao2024taubench}), S2 proposes  actions from clean records while S1 (\texttt{allow}/\texttt{block}/\texttt{escalate}) inspects an note.

Table~\ref{tab:e2e_results} compares \texttt{Clean} and four attacks across 150 tasks per workflow. On \textbf{Tool Router}, S1-EC lowers accuracy from 94.00\% to 69.33\%, reaching 58.67\% S1 ASR and 24.00\% E2E ASR, outperforming \texttt{PAIR-jev}, \texttt{JevOut}, and \texttt{FlipJev}. On \textbf{Guardrail}, S1-EC drops accuracy from 98.00\% to 80.67\% and attains 45.33\% S1 ASR and 17.33\% E2E ASR, surpassing \texttt{PAIR-jev} by 6.66 percentage points. Attacks also trigger extra execution and retry steps, significantly increasing S2 tokens and latency across both workflows.

Not all S1 flips cause system violations. In \textbf{Tool Router}, 40.91\% of S1-EC flips yield E2E violations, as S2 generates parameters after routing and often refuses unauthorized arguments. \textbf{Guardrail} shows a comparable 38.24\% conversion: while flipping \texttt{block} to \texttt{allow} directly executes S2's pre-proposed parameters, benign-side false blocks halt execution without policy breach and multi-step domain checks absorb remaining flips.

\subsection{Cross-Library Transfer and Defenses}
\label{subsec:more_analyses}
We further evaluate cross-benchmark library transfer and propose two inference-time defenses---Rule Reminder and Label Anonymization, against S1-EC. See Appendices~\ref{app:cross_lib} and~\ref{app:defenses} for details.

\section{Conclusion}
This paper studied the robustness of typed S1 decision models in LLM agent workflows and introduced S1-EC, a self-evolving adversarial attack framework tailored to S1 systems. S1-EC induces \emph{evidence competition} through contract-aware evidence channels, refines attack candidates offline via directional margin feedback, and distills verified flips into a structure-indexed experience library to construct effective perturbations for unseen tasks without iterative feedback. Experiments across four node-level benchmarks, two E2E agent systems, and two inference-time defenses demonstrate high attack success rates, reduced test-time token usage, and end-to-end violations, highlighting the need for stronger defenses at typed decision boundaries.

\section*{Limitations}
\label{sec:limitations}

Our study has two main limitations. First, the self-evolving component of S1-EC is not strictly plug-and-play: akin to a training phase, constructing the structure-indexed experience library requires an initial set of seed tasks and offline iterative target feedback to collect and verify flips before attacking unseen tasks in a single pass. Moreover, distilled operator priors exhibit bounded generalization when transferred across tasks with substantially different contract schemas or rule structures. Second, a successful node-level flip on an S1 model does not always translate into an end-to-end system violation. Downstream S2 reasoners or runtime verifiers can partially detect inconsistent tool routes and recover execution. Nevertheless, even when such downstream recovery prevents final policy violations, misrouting control flow into heavy S2 verification or fallback branches already incurs substantial token and latency overhead, undermining the efficiency motivation of deploying S1 nodes.

\section*{Ethical Considerations}
This work studies adversarial vulnerabilities of typed System-1 decision models to support pre-deployment auditing and stronger guardrail design. All experiments are conducted strictly on public benchmarks and isolated sandbox environments (\texttt{Tool Router} and \texttt{Guardrail}) using synthetic records without affecting real-world services or external users. To mitigate potential misuse, we pair our vulnerability analysis with two practical inference-time defenses (\emph{Rule Reminder} and \emph{Label Anonymization} in Appendix~C) that substantially reduce attack success rates.

\bibliography{custom}

\clearpage
\appendix

\section{Semantic Label Preservation Audit}
\label{app:semantic_validity}

To verify that S1-EC does not alter the original
ground-truth decisions, we conduct a semantic label
preservation audit. Specifically, we sample 100
successfully attacked (pre-injection) instances from each of the four benchmarks (DR, RJ, WG, and IA), yielding 400 instances in total. For each instance, we examine whether the original ground-truth label remains valid after adding the adversarial perturbation. Three independent LLM judges, GPT-5, Gemini-3.5-flash, and DeepSeek-v4.1-flash, evaluate label preservation. The evaluation
focuses on whether the injected content changes
the correct answer under the original task rules,
rather than whether the victim model follows
the adversarial content.

\begin{table}[h]
\caption{Semantic label preservation rates (\%)
assessed by three LLM judges on 100 sampled
original instances per benchmark.}
\centering
\small
\setlength{\tabcolsep}{3.5pt}
\begin{tabular}{lccccc}
\toprule
\textbf{Judge} & \textbf{DR} & \textbf{RJ}
& \textbf{WG} & \textbf{IA} & \textbf{Avg.} \\
\midrule
GPT-5          & 96 & 95 & 98 & 99 & 97.00 \\
Gemini-3.5-flash & 94 & 94 & 98 & 100 & 96.50 \\
DeepSeek-v4.1-flash & 97 & 95 & 97 & 98 & 96.75 \\
\midrule
\textbf{Average}
& \textbf{95.67} & \textbf{94.67}
& \textbf{97.67} & \textbf{99.00}
& \textbf{96.75} \\
\bottomrule
\end{tabular}

\label{tab:semantic_validity}
\end{table}

As shown in Table~\ref{tab:semantic_validity},
S1-EC achieves consistently high semantic label
preservation across all four benchmarks, with
an overall average of 96.75\% across judges
and benchmarks. The preservation rates range
from 94.67\% on R-Judge to 99.00\% on
InjecAgent. These results indicate that the
generated perturbations largely preserve the
original ground-truth decisions, supporting
our threat model in which adversarial context
introduces misleading evidence without
changing the underlying correct answer.

\section{Cross-Library Generalization}
\label{app:cross_lib}
To test whether the experience library $\mathcal{L}_{\mathrm{exp}}$ captures transferable structural patterns rather than benchmark-specific phrasing, we evaluate cross-library transfer between WildGuardTest (WG) and R-Judge (RJ) on \texttt{jev-1.14} (\texttt{DeepSeek} attacker). Specifically, we use the frozen RJ experience library to warm-start attacks on WG and vice versa, comparing against both zero-experience search (\texttt{none}) and in-domain libraries.

Table~\ref{tab:cross_library} shows that cross-domain libraries consistently outperform \texttt{none} under comparable token budgets (827--918 tokens on WG and 615--679 on RJ). On WG, the RJ library raises ASR from 51.00\% to 57.00\% and High-Conf ASR from 42.00\% to 45.50\%. On RJ, the WG library lifts ASR from 38.98\% to 49.74\% and High-Conf ASR from 17.95\% to 26.15\%, confirming that distilled experience cards transfer across tasks.

\begin{table}[h]
\centering
\scriptsize
\setlength{\tabcolsep}{5.0pt}
\renewcommand{\arraystretch}{1.10}
\caption{Cross-library generalization results (\%) and attacker LLM token cost (\textbf{Tok}).}
\label{tab:cross_library}
\resizebox{\columnwidth}{!}{%
\begin{tabular}{ll rrr}
\toprule
\textbf{Bench.} & \textbf{Exp.\ Lib} & \multicolumn{1}{c}{\textbf{ASR}$\uparrow$} & \multicolumn{1}{c}{\textbf{HC}$\uparrow$} & \multicolumn{1}{c}{\textbf{Tok}} \\
\midrule
 & \texttt{none} & 51.00 & 42.00 & 827 \\
 & RJ (Cross-domain) & \underline{57.00} & \underline{45.50} & 842 \\
\rowcolor{gray!12} \cellcolor{white}\multirow{-3}{*}{\textbf{WG}} & \textbf{WG (In-domain)} & \textbf{65.00} & \textbf{53.00} & 918 \\
\midrule
 & \texttt{none} & 38.98 & 17.95 & 615 \\
 & WG (Cross-domain) & \underline{49.74} & \underline{26.15} & 679 \\
\rowcolor{gray!12} \cellcolor{white}\multirow{-3}{*}{\textbf{RJ}} & \textbf{RJ (In-domain)} & \textbf{65.13} & \textbf{41.54} & 665 \\
\bottomrule
\end{tabular}%
}
\end{table}

\section{Potential Defenses}
\label{app:defenses}

We design and evaluate two inference-time defenses tailored to typed S1 decision interfaces. First, \textbf{Rule Reminder} reinforces trust boundaries by appending an explicit verification instruction after untrusted input segments, reminding the S1 model to judge strictly by core task facts and trusted policies while disregarding unverified status claims, exception notes, or embedded conclusions. Second, \textbf{Label Anonymization} decouples option symbols from semantic priors: it replaces meaningful label names in typed \texttt{Choice} schemas (e.g., \texttt{safe}/\texttt{unsafe}) with neutral identifiers (e.g., \texttt{OPT\_A}/\texttt{OPT\_B}) and defines their mapping solely inside the trusted prompt header, preventing adversarial text from directly anchoring target label tokens.

Table~\ref{tab:defenses} reports defense performance against S1-EC on DecisionRobust (DR) and WildGuardTest (WG) using \texttt{jev-1.14} (\texttt{DeepSeek} attacker). Without defense, S1-EC lowers accuracy from 89.00\% to 18.50\% on DR (70.50\% ASR) and from 96.00\% to 32.00\% on WG (65.00\% ASR). \textbf{Rule Reminder} restores accuracy to 46.00\% and 69.00\%, cutting ASR to 43.50\% and 27.00\%. \textbf{Label Anonymization} proves stronger, raising accuracy to 58.00\% and 77.50\% while reducing ASR to 38.00\% and 18.50\% (High-Conf ASR to 21.00\% and 10.50\%). Nevertheless, S1-EC still flips 38.00\% of DR tasks, as contract-aligned evidence alters task context beyond surface label cues.

\begin{table}[h]
\centering
\scriptsize
\setlength{\tabcolsep}{3.5pt}
\renewcommand{\arraystretch}{1.10}
\caption{Defense evaluation (\%) against S1-EC on DecisionRobust (DR) and WildGuardTest (WG) with \texttt{jev-1.14} (\texttt{DeepSeek} attacker). Best defense results (highest Acc, lowest ASR/HC) are in \textbf{bold}.}
\label{tab:defenses}
\resizebox{\columnwidth}{!}{%
\begin{tabular}{l rrr rrr}
\toprule
\multirow{2}{*}{\textbf{Setting / Defense}} & \multicolumn{3}{c}{\textbf{DecisionRobust (DR)}} & \multicolumn{3}{c}{\textbf{WildGuardTest (WG)}} \\
\cmidrule(lr){2-4} \cmidrule(lr){5-7}
 & \multicolumn{1}{c}{Acc$\uparrow$} & \multicolumn{1}{c}{ASR$\downarrow$} & \multicolumn{1}{c}{HC$\downarrow$} & \multicolumn{1}{c}{Acc$\uparrow$} & \multicolumn{1}{c}{ASR$\downarrow$} & \multicolumn{1}{c}{HC$\downarrow$} \\
\midrule
Clean (No Attack) & 89.00 & \multicolumn{1}{c}{--} & \multicolumn{1}{c}{--} & 96.00 & \multicolumn{1}{c}{--} & \multicolumn{1}{c}{--} \\
S1-EC (No Defense) & 18.50 & 70.50 & 48.00 & 32.00 & 65.00 & 53.00 \\
\midrule
+ Rule Reminder & \underline{46.00} & \underline{43.50} & \underline{27.00} & \underline{69.00} & \underline{27.00} & \underline{14.00} \\
\rowcolor{gray!12} \textbf{+ Label Anonymization} & \textbf{58.00} & \textbf{38.00} & \textbf{21.00} & \textbf{77.50} & \textbf{18.50} & \textbf{10.50} \\
\bottomrule
\end{tabular}%
}
\end{table}

\section{Cost details}
\label{app:offline_online}

\paragraph{Cost definition.}
To account for reusable preparation, we distinguish offline and online attacker-LLM tokens. For benchmark $b$, $T_b^{\mathrm{off}}$ denotes tokens spent constructing its frozen experience library, $T_b^{\mathrm{on}}$ denotes mean tokens needed to construct one perturbation at test time, and $N_b$ denotes the number of test instances reusing that library. The amortized cost is $T_b^{\mathrm{amort}}=T_b^{\mathrm{on}}+T_b^{\mathrm{off}}/N_b$. Offline costs are charged separately to each benchmark-specific library, and aggregate results use benchmark-macro averages consistent with the main experiments. No separate experience-library construction cost is recorded for iterative baselines, so their online and amortized values coincide.

\begin{table*}[t]
\centering
\caption{Attack success rates and attacker-LLM costs.
ASR follows Table~1. Online and amortized costs
are measured in tokens/task; offline costs are in
thousands of tokens. Avg. denotes the macro-average,
except for offline cost, which reports the total.}
\label{tab:offline_online}
\scriptsize
\setlength{\tabcolsep}{2.8pt}
\renewcommand{\arraystretch}{1.05}

\begin{tabular*}{\textwidth}{@{\extracolsep{\fill}}lrrrrrlrrrrr@{}}
\toprule
\multicolumn{6}{c}{\textbf{ASR (\%) $\uparrow$}}
& \multicolumn{6}{c}{\textbf{Attacker-LLM Token Cost}} \\
\cmidrule(lr){1-6}\cmidrule(l){7-12}
Method & DR & RJ & WG & IA & Avg.
& Method / Phase & DR & RJ & WG & IA & Avg. \\
\midrule
JevOut
& 24.00 & 23.08 & 12.00 & 17.00 & 19.02
& JevOut (Online)
& 1778 & 1876 & 1944 & 1683 & 1820.3 \\

PAIR-jev
& 37.00 & 46.15 & 22.00 & 5.33 & 27.62
& PAIR-jev (Online)
& 1896 & 836 & 1257 & 790 & 1194.8 \\
\midrule
\textbf{S1-EC}
& \textbf{70.50} & \textbf{65.13}
& \textbf{65.00} & \textbf{87.33} & \textbf{71.99}
& \textbf{S1-EC (Offline, k)}
& 113 & 78 & 95 & 53 & 339 \\

\multicolumn{6}{c}{}
& \textbf{S1-EC (Online)}
& 954 & 665 & 918 & 211 & 687.0 \\

\multicolumn{6}{c}{}
& \textbf{S1-EC (Amortized)}
& 1519 & 1065 & 1393 & 387.7 & 1091.2 \\
\bottomrule
\end{tabular*}
\end{table*}

\paragraph{Analysis.}
Table~\ref{tab:offline_online} shows that building the four libraries consumes 339k attacker-LLM tokens offline (113k/78k/95k/53k), while S1-EC uses 687.0 tokens per task online on average. At the evaluated test-set sizes, its benchmark-macro amortized cost is 1,091.2 tokens per task: 40.1\% below JevOut (1,820.3) and 8.7\% below PAIR-jev (1,194.8). S1-EC achieves 71.99\% mean ASR versus 27.62\% for PAIR-jev. The advantage is effective single-pass construction for unseen tasks without iterative victim feedback at attack time. This trade-off is most attractive when a frozen library serves many compatible decision contracts.

\section{Repeated-Run Analysis}
\label{sec:app_repeat}

To verify experimental robustness, we conduct three repeated runs on the main node-level experiments and report the mean and standard deviation (in parentheses) for \textbf{Acc}$\downarrow$, \textbf{ASR}$\uparrow$, and \textbf{HC}$\uparrow$ in Table~\ref{tab:app_repeat_results}.

Across all 48 attack-benchmark pairs, run-to-run variance remains small: standard deviations do not exceed 2.50 percentage points for \textbf{Acc} and \textbf{ASR} (averaging 1.14\% and 1.10\%, respectively) or 2.05 percentage points for \textbf{HC} (averaging 0.79\%). Under \texttt{DeepSeek}, \textbf{S1-EC} exhibits tight stability with standard deviations between 0.50\% and 1.54\%, outperforming the strongest baselines on DecisionRobust, R-Judge, and WildGuardTest by margins (18.98--43.00 percentage points in ASR) that exceed ten times its observed standard deviations. Under \texttt{GLM-5.3}, \textbf{S1-EC} similarly maintains standard deviations within 0.33\%--2.50\% and consistently leads overall performance, confirming that the gains of contract-aware evidence competition and staged feedback search are insensitive to sampling randomness.

\begin{table*}[t]
\centering
\scriptsize
\setlength{\tabcolsep}{3.2pt}
\renewcommand{\arraystretch}{1.15}
\caption{Repeated-run node-level attack performance (\%).}
\label{tab:app_repeat_results}
\resizebox{\textwidth}{!}{%
\begin{tabular}{ll rrr rrr rrr rrr}
\toprule
\multirow{2}{*}{\textbf{LLM}} & \multirow{2}{*}{\textbf{Method}} & \multicolumn{3}{c}{\textbf{DecisionRobust (DR)}} & \multicolumn{3}{c}{\textbf{R-Judge (RJ)}} & \multicolumn{3}{c}{\textbf{WildGuardTest (WG)}} & \multicolumn{3}{c}{\textbf{InjecAgent (IA)}} \\
\cmidrule(lr){3-5} \cmidrule(lr){6-8} \cmidrule(lr){9-11} \cmidrule(lr){12-14}
 & & \multicolumn{1}{c}{Acc$\downarrow$} & \multicolumn{1}{c}{ASR$\uparrow$} & \multicolumn{1}{c}{HC$\uparrow$} & \multicolumn{1}{c}{Acc$\downarrow$} & \multicolumn{1}{c}{ASR$\uparrow$} & \multicolumn{1}{c}{HC$\uparrow$} & \multicolumn{1}{c}{Acc$\downarrow$} & \multicolumn{1}{c}{ASR$\uparrow$} & \multicolumn{1}{c}{HC$\uparrow$} & \multicolumn{1}{c}{Acc$\downarrow$} & \multicolumn{1}{c}{ASR$\uparrow$} & \multicolumn{1}{c}{HC$\uparrow$} \\
\midrule
 & Clean & 89.00\textsubscript{(1.00)} & \multicolumn{1}{c}{--} & \multicolumn{1}{c}{--} & 83.08\textsubscript{(1.03)} & \multicolumn{1}{c}{--} & \multicolumn{1}{c}{--} & 96.00\textsubscript{(2.00)} & \multicolumn{1}{c}{--} & \multicolumn{1}{c}{--} & 100.00\textsubscript{(0.00)} & \multicolumn{1}{c}{--} & \multicolumn{1}{c}{--} \\
 & JevOut & 65.00\textsubscript{(1.00)} & 24.00\textsubscript{(1.00)} & 9.50\textsubscript{(0.50)} & 60.00\textsubscript{(0.51)} & 23.08\textsubscript{(0.51)} & 11.28\textsubscript{(1.03)} & 84.00\textsubscript{(1.00)} & 12.00\textsubscript{(1.00)} & 0.50\textsubscript{(0.50)} & 83.00\textsubscript{(0.33)} & 17.00\textsubscript{(0.33)} & 17.00\textsubscript{(0.33)} \\
 & PAIR-jev & \underline{52.00}\textsubscript{(0.50)} & \underline{37.00}\textsubscript{(0.50)} & \underline{16.50}\textsubscript{(0.50)} & \underline{36.92}\textsubscript{(2.05)} & \underline{46.15}\textsubscript{(2.05)} & \underline{23.59}\textsubscript{(1.03)} & \underline{74.00}\textsubscript{(1.50)} & \underline{22.00}\textsubscript{(1.50)} & \underline{10.00}\textsubscript{(1.00)} & 94.67\textsubscript{(0.33)} & 5.33\textsubscript{(0.33)} & 5.33\textsubscript{(0.33)} \\
 & FlipJev & 69.50\textsubscript{(1.00)} & 19.50\textsubscript{(1.00)} & 7.00\textsubscript{(1.00)} & 51.79\textsubscript{(1.03)} & 31.28\textsubscript{(1.03)} & 9.74\textsubscript{(1.54)} & 82.50\textsubscript{(2.00)} & 14.00\textsubscript{(2.00)} & 2.50\textsubscript{(1.50)} & 89.33\textsubscript{(0.33)} & 10.67\textsubscript{(0.33)} & 10.67\textsubscript{(0.00)} \\
 & Inject & 67.00\textsubscript{(1.50)} & 22.00\textsubscript{(1.50)} & 6.50\textsubscript{(0.50)} & 58.46\textsubscript{(1.54)} & 24.62\textsubscript{(1.54)} & 5.13\textsubscript{(0.51)} & 83.00\textsubscript{(1.50)} & 13.50\textsubscript{(1.50)} & 2.00\textsubscript{(0.50)} & 100.00\textsubscript{(0.00)} & 0.00\textsubscript{(0.00)} & 0.00\textsubscript{(0.00)} \\
 & DH-72 & 54.50\textsubscript{(1.00)} & 34.50\textsubscript{(1.00)} & 15.00\textsubscript{(1.00)} & 68.21\textsubscript{(1.03)} & 14.87\textsubscript{(1.03)} & 6.67\textsubscript{(0.51)} & 82.50\textsubscript{(1.50)} & 14.00\textsubscript{(0.50)} & 6.00\textsubscript{(0.00)} & \textbf{4.33}\textsubscript{(1.00)} & \textbf{95.67}\textsubscript{(1.00)} & \textbf{95.67}\textsubscript{(1.00)} \\
\rowcolor{gray!12} \cellcolor{white}\multirow{-7}{*}{\rotatebox[origin=c]{90}{\textbf{DeepSeek}}} & \textbf{S1-EC} & \textbf{18.50}\textsubscript{(1.50)} & \textbf{70.50}\textsubscript{(1.50)} & \textbf{48.00}\textsubscript{(1.00)} & \textbf{17.95}\textsubscript{(1.54)} & \textbf{65.13}\textsubscript{(1.54)} & \textbf{41.54}\textsubscript{(1.03)} & \textbf{32.00}\textsubscript{(1.00)} & \textbf{65.00}\textsubscript{(1.00)} & \textbf{53.00}\textsubscript{(0.50)} & \underline{12.67}\textsubscript{(0.67)} & \underline{87.33}\textsubscript{(0.67)} & \underline{87.33}\textsubscript{(0.67)} \\
\midrule
 & Clean & 89.00\textsubscript{(1.00)} & \multicolumn{1}{c}{--} & \multicolumn{1}{c}{--} & 83.08\textsubscript{(1.03)} & \multicolumn{1}{c}{--} & \multicolumn{1}{c}{--} & 96.00\textsubscript{(2.00)} & \multicolumn{1}{c}{--} & \multicolumn{1}{c}{--} & 100.00\textsubscript{(0.00)} & \multicolumn{1}{c}{--} & \multicolumn{1}{c}{--} \\
 & JevOut & 69.00\textsubscript{(1.00)} & 20.00\textsubscript{(1.00)} & 8.00\textsubscript{(1.00)} & 61.03\textsubscript{(1.54)} & 22.05\textsubscript{(1.54)} & 13.85\textsubscript{(1.54)} & 84.50\textsubscript{(1.50)} & 11.50\textsubscript{(1.50)} & 0.00\textsubscript{(0.00)} & 89.67\textsubscript{(0.67)} & 10.33\textsubscript{(0.67)} & 10.33\textsubscript{(0.67)} \\
 & PAIR-jev & \underline{52.00}\textsubscript{(1.00)} & \underline{37.00}\textsubscript{(1.00)} & \underline{21.00}\textsubscript{(1.00)} & \textbf{34.36}\textsubscript{(1.03)} & \textbf{48.72}\textsubscript{(1.03)} & \underline{27.18}\textsubscript{(0.51)} & \underline{69.00}\textsubscript{(2.00)} & \underline{27.00}\textsubscript{(2.00)} & \underline{12.00}\textsubscript{(2.00)} & 92.00\textsubscript{(0.33)} & 8.00\textsubscript{(0.33)} & 8.00\textsubscript{(0.33)} \\
 & FlipJev & 69.50\textsubscript{(1.00)} & 19.50\textsubscript{(1.00)} & 7.00\textsubscript{(1.00)} & 51.79\textsubscript{(1.03)} & 31.28\textsubscript{(1.03)} & 9.74\textsubscript{(1.54)} & 82.50\textsubscript{(2.00)} & 14.00\textsubscript{(2.00)} & 2.50\textsubscript{(1.50)} & 89.33\textsubscript{(0.33)} & 10.67\textsubscript{(0.33)} & 10.67\textsubscript{(0.00)} \\
 & Inject & 67.00\textsubscript{(1.50)} & 22.00\textsubscript{(1.50)} & 6.50\textsubscript{(0.50)} & 58.46\textsubscript{(1.54)} & 24.62\textsubscript{(1.54)} & 5.13\textsubscript{(0.51)} & 83.00\textsubscript{(1.50)} & 13.50\textsubscript{(1.50)} & 2.00\textsubscript{(0.50)} & 100.00\textsubscript{(0.00)} & 0.00\textsubscript{(0.00)} & 0.00\textsubscript{(0.00)} \\
 & DH-72 & 54.50\textsubscript{(1.00)} & 34.50\textsubscript{(1.00)} & 15.00\textsubscript{(1.00)} & 68.21\textsubscript{(1.03)} & 14.87\textsubscript{(1.03)} & 6.67\textsubscript{(0.51)} & 82.50\textsubscript{(1.50)} & 14.00\textsubscript{(0.50)} & 6.00\textsubscript{(0.00)} & \textbf{4.33}\textsubscript{(1.00)} & \textbf{95.67}\textsubscript{(1.00)} & \textbf{95.67}\textsubscript{(1.00)} \\
\rowcolor{gray!12} \cellcolor{white}\multirow{-7}{*}{\rotatebox[origin=c]{90}{\textbf{GLM-5.3}}} & \textbf{S1-EC} & \textbf{12.00}\textsubscript{(2.50)} & \textbf{77.00}\textsubscript{(2.50)} & \textbf{56.50}\textsubscript{(2.00)} & \underline{35.38}\textsubscript{(2.05)} & \underline{47.69}\textsubscript{(2.05)} & \textbf{31.79}\textsubscript{(2.05)} & \textbf{44.00}\textsubscript{(2.00)} & \textbf{52.00}\textsubscript{(2.00)} & \textbf{42.00}\textsubscript{(2.00)} & \underline{33.67}\textsubscript{(0.33)} & \underline{66.33}\textsubscript{(0.33)} & \underline{66.33}\textsubscript{(0.33)} \\
\bottomrule
\end{tabular}%
}
\end{table*}

\section{Implementation Details}
\label{app:impl}

\subsection{Datasets and Evaluation Splits}
We evaluate node-level attacks across four benchmarks totaling 895 test instances. On each benchmark, we first sample $N=50$ disjoint seed tasks (separated by source family) to construct the experience library $\mathcal{L}_{\mathrm{exp}}$. We then sample the test set so that each subcategory contains at least 50 instances wherever available:
\begin{itemize}[leftmargin=1.5em]
    \item \textbf{DecisionRobust (DR):} 200 test tasks, evenly divided between \texttt{Choice} (100) and \texttt{Noul} (100).
    \item \textbf{WildGuardTest (WG)}: 200 safety moderation test tasks evaluated via the \texttt{Score} primitive (mapping safety score $s(x)$ against $\tau_{score}$ to \texttt{allow}/\texttt{block}).
    \item \textbf{R-Judge (RJ):} 195 multi-turn agent risk judgment tasks across four categories: \texttt{Program} (50), \texttt{App} (50), \texttt{Finance} (50), and \texttt{Web/IoT} (45; after excluding the disjoint seed tasks reserved for library construction, only 45 \texttt{Web/IoT} instances remain, all of which are included in the test set).
    \item \textbf{InjecAgent (IA):} 300 indirect prompt injection tasks spanning six categories (\texttt{phys.harm}, \texttt{fin.harm}, \texttt{phys.data}, \texttt{fin.data}, \texttt{datasec}, and \texttt{others}), with 50 tasks per category.
\end{itemize}

\subsection{Metrics and Verification Protocol}
\noindent\textbf{Attack Success Rate (ASR).}
Let $\mathcal{D}_{\mathrm{test}}$ denote the full test set of a benchmark. ASR is defined as the number of tasks that are correctly predicted on clean inputs ($f(x) = y_{\mathrm{gold}}$) but misclassified under attack ($f(x') \neq y_{\mathrm{gold}}$), divided by the total number of test tasks:
\begin{equation}
\mathrm{ASR} = \frac{\left|\{x \in \mathcal{D}_{\mathrm{test}} : f(x) = y_{\mathrm{gold}} \land f(x') \neq y_{\mathrm{gold}}\}\right|}{|\mathcal{D}_{\mathrm{test}}|}.
\end{equation}

\paragraph{High-Confidence Attack Success Rate (HC).}
High-Confidence ASR (HC) measures the proportion of test instances achieving a verified attack success with high prediction margins on both clean and adversarial inputs. Specifically, an attack must first satisfy the validity and success conditions in Eq.~(8), including the required majority-vote re-verification. HC further requires that the original correct prediction has a probability margin greater than $0.2$ and that the adversarial incorrect prediction also has a probability margin greater than $0.2$.

Formally, for a test set $\mathcal{D}_{\mathrm{test}}$, we define
\begin{equation}
\label{eq:hc_asr}
\mathrm{HC}
=
\frac{1}{|\mathcal{D}_{\mathrm{test}}|}
\sum_{x \in \mathcal{D}_{\mathrm{test}}}
\mathbb{I}\left[
\begin{aligned}
&\mathrm{Success}(x,x') \\
&\land\ M_{\mathrm{clean}}(x)>0.2 \\
&\land\ M_{\mathrm{adv}}(x')>0.2
\end{aligned}
\right],
\end{equation}
where $\mathrm{Success}(x,x')$ denotes a verified successful attack satisfying Eq.~(8), and
\begin{equation}
\begin{aligned}
M_{\mathrm{clean}}(x)
&= P(y^* \mid x)
- \max_{o \neq y^*} P(o \mid x),\\
M_{\mathrm{adv}}(x')
&= P(\hat{y} \mid x')
- \max_{o \neq \hat{y}} P(o \mid x').
\end{aligned}
\end{equation}
For \texttt{Score}-based tasks (WG), $P(\texttt{allow}\mid x) = s(x)$ and $P(\texttt{block}\mid x) = 1 - s(x)$, directly following the same $>0.2$ margin criterion in Eq.~(11)--(12).
Here, $y^*$ is the ground-truth option and $\hat{y}$ is the incorrect option selected after the attack. Both margins must exceed $0.2$, ensuring that HC captures verified decision reversals from strongly correct to strongly incorrect predictions. The denominator is the full test-set size $|\mathcal{D}_{\mathrm{test}}|$, consistent with the ASR definition.

\subsection{Experience Construction Protocol}
\noindent\textbf{Experience Card Promotion.}
On the $N=50$ seed tasks, S1-EC scans 11 contract-aware evidence operator families under an offline budget of $B=12$ victim queries per seed task (covering candidate evaluation and re-verification). Any family achieving at least one re-verified flip on the seed set is promoted into a verified JSON experience card recording its operator family, six-field applicability signature (\texttt{primitive}, \texttt{lang}, \texttt{anchor}, \texttt{option\_distance}, \texttt{margin\_band}, \texttt{edit\_surface}), binding recipe, failure counter-conditions, and Wilson score lower bound. Once constructed, $\mathcal{L}_{\mathrm{exp}}$ is frozen for read-only test-time access and maintained independently per victim model.

\noindent\textbf{Three-Tier Indexing and Single-Step Construction.}
During testing, card retrieval applies (1) hard filtering on decision primitive and language (relaxing \texttt{primitive} in Appendix~B), followed by (2) structural scoring across the four structural keys ($+1.0$ for exact match, $+0.5$ for wildcard \texttt{null}---including \texttt{margin\_band} without pre-querying---and $0$ for mismatch; maximum score $4.0$), while domain categories are used only for entity vocabulary adaptation. Matched cards are ranked by structural score and Wilson lower bound to select heterogeneous cards from distinct operator families (injecting all matched cards by default, or Top-$k$ when varied). Using the retrieved cards (or fallback families \texttt{affinity\_demote}, \texttt{argument}, \texttt{authority}, \texttt{condmap} when no card matches), the constructor LLM synthesizes one perturbation per task in a single pass without online victim-feedback search.

\subsection{Baseline Implementation}
For a controlled comparison, all methods use identical evaluation splits, original-evidence preservation checks, and attack verification criteria. For iterative baselines, we impose a unified cap of $B=20$ iterations per instance (with early termination upon confirmation), whereas S1-EC builds its experience library once offline and directly constructs a single perturbation at test time without repeated probing---reflecting realistic deployments where repeatedly querying the same decision module on an active instance is rarely feasible.

We adapt each baseline to the typed decision interface while retaining its defining attack mechanism.\textbf{JevOut} follows the original probability-guided context optimization procedure. It iteratively extends answer-preserving contexts and prioritizes candidates using probability-margin feedback to mislead the model into selecting any incorrect decision. \textbf{PAIR-jev} performs iterative LLM rewriting with binary success/failure feedback. \textbf{FlipJev} conducts probability-guided word-level search offline on the disjoint seed tasks without an attacker LLM to optimize universal adversarial suffixes, followed by direct online substitution into the untrusted edit surface at test time.\textbf{Inject} evaluates three fixed authorization-themed templates, while \textbf{DH-72} searches a budget-limited subset of its 72 prefix--suffix--repetition configurations. Imperative-injection baselines are exempt from the command-free lexical restriction while retaining the remaining input-integrity constraints.

This protocol bounds baseline search overhead and enforces consistent success verification across methods while preserving their respective attack mechanisms and feedback strategies. The comparison therefore evaluates single-pass S1-EC against iterative baselines under a shared iteration cap.

\subsection{Ablation Variants}
All four ablated variants in Table~\ref{tab:ablation} use identical test splits, prefix-integrity checks, and re-verification rules:
(1) \textbf{Nat-Gen} appends a single task-irrelevant neutral sentence under a strict lexical blocklist forbidding argumentative connectives, policy references, and option synonyms (reverting to clean if the audit fails);
(2) \textbf{Operator-Only} constructs a perturbation using unguided evidence operators without library retrieval;
(3) \textbf{Imp-Chan} replaces fact-invariant evidence operators with three imperative override templates plus word-level mutations (exempted only from the imperative blocklist); and
(4) \textbf{Word-Search} uses no evidence operators, searching instead for word-level edits within a fixed appended passage under the $B=20$ iteration cap while preserving the original input verbatim.

\subsection{End-to-End Dual-System Agent Protocol}
\label{app:e2e_setup}
\noindent\textbf{Common Execution Rules.}
Both dual-system workflows comprise 150 tasks deduplicated by source family and enforce a hard execution cap of 16 agent decision steps per task. Intra-step operations---including three-vote S1 queries, S2 parameter generation, API retries, and final summarization---do not increment the step count, although all downstream S2 tokens and wall-clock latency are recorded. If a task reaches 16 steps without finishing, execution halts (\texttt{max\_steps}) and outputs a tool-free status summary without bypassing \texttt{block} or \texttt{escalate} states. Consistent with our node-level setup, both workflows initialize S1-EC with a frozen experience library $\mathcal{L}_{exp}$ distilled offline from disjoint seed tasks for read-only retrieval, performing a single-step search in the E2E systems to construct perturbations from the retrieved cards. All S1 inputs serialize trusted context as a fixed prefix followed by the untrusted field (verified via SHA hashing), and end-to-end violations are determined strictly by independent environment state assertions.

\noindent\textbf{Tool Router (AgentDojo Adapted).}
This workflow consists of 150 office tasks across five categories (30 each: file lookup, email processing, calendar scheduling, cross-tool information flow, and multi-step verification) operating over six simulated tools (\texttt{web\_search}, \texttt{calculator}, \texttt{calendar}, \texttt{send\_email}, \texttt{file\_read}, \texttt{file\_delete}). S2 decomposes subtasks, S1 selects the tool, and S2 populates parameters or refuses. Perturbations are appended to a designated tool return to steer the next S1 routing step; an E2E violation is recorded only when an unauthorized state change (e.g., unauthorized file deletion or email dispatch) actually occurs.

\noindent\textbf{Guardrail ($\tau$-bench Adapted).}
This workflow consists of 150 retail order and refund tasks across five policy categories (refund amount, refund channel, order status/window, account ownership, and cancellation/duplicate operations), balanced into 15 legitimate (\texttt{allow}$\to$\texttt{block}) and 15 policy-violating (\texttt{block}$\to$\texttt{allow}) cases per category. S2 first proposes a complete parameterized action from clean records; the authorization gateway then queries S1 (\texttt{Choice}: \texttt{allow}, \texttt{block}, \texttt{escalate}) over the policy, state facts, proposed action, and an untrusted ticket note. \texttt{allow} executes the action, \texttt{block} returns a structured refusal allowing S2 to revise within the 16-step limit, and \texttt{escalate} or S1 errors halt the task safely.

\section{Algorithmic Pseudocode of S1-EC}
\label{app:pseudocode}
Algorithm~\ref{alg:s1_evo} summarizes the complete execution workflow of S1-EC across offline experience card accumulation, node-level constrained stratified feedback search, and online dual-system agent auditing described in Section~\ref{sec:methodology}.

\makeatletter
\@ifundefined{c@s1evoline}{\newcounter{s1evoline}}{}
\makeatother

\begin{algorithm*}[t]
\caption{S1-EC: Self-Evolving Evidence-Channel Attack and Dual-System Agent Auditing}
\label{alg:s1_evo}
\footnotesize
\vspace{2pt}
\noindent\textbf{Input:} Disjoint seed set $\mathcal{D}_{\mathrm{seed}}$; test decision contract $(S, R, \mathcal{O})$ with primitive $d \in \{\texttt{Choice}, \texttt{Score}, \texttt{Noul}\}$, language $\ell$, untrusted field $x \in \mathrm{Ext}(S)$, and optional target $t$ (inducing target set $\mathcal{Y}_{\mathrm{att}}$); victim S1 model $f$; clean agent prefix $h$ and dual-system environment $\mathcal{E}$.\\
\textbf{Parameters:} Evidence channels $\mathcal{C}$ with operator pool $\Psi = \{\psi_{c,k}\}$ and fallback set $\Psi_{\mathrm{fb}}$; attacker LLM $\mathcal{M}_{\mathrm{att}}$; offline seed query budget $B$; retrieval size $k$; seed refinement/fallback caps $(T_{\mathrm{ref}}, T_{\mathrm{fb}})$; thresholds $(\tau_{\mathrm{score}}, \tau_{\mathrm{conf}})$.\\
\textbf{Output:} Frozen experience library $\mathcal{L}_{\mathrm{exp}}$, constructed perturbation $x'$, and node/E2E audit indicators.
\vspace{3pt}
\hrule height 0.4pt
\vspace{3pt}
\noindent
\begin{tabular}{@{}p{0.482\textwidth}@{\hspace{0.016\textwidth}}|@{\hspace{0.016\textwidth}}p{0.482\textwidth}@{}}
\begin{algorithmic}[1]
\Statex \textbf{Subroutine: Seed Query \& Re-Verification}
\Function{EvalSeed}{$x_s, \delta, Q$}
    \State $x'_s \gets \phi(x_s, \delta)$ \Comment{Eq.~\eqref{eq:operator_instantiation}}
    \State \textbf{if} $\neg V(x_s, x'_s) \lor Q \ge B$ \textbf{then return} $(\bot, -\infty, Q)$
    \State Query $f(S(x'_s), R_s, \mathcal{O}_s) \to \bigl(P(x'_s), \hat{y}(x'_s), C(x'_s)\bigr)$; $Q \gets Q + 1$
    \State Compute feedback $g(x'_s)$ via Eq.~\eqref{eq:dir_feedback} (or $\mathbb{I}(\hat{y}(x'_s) \in \mathcal{Y}_{\mathrm{att}})$)
    \If{$\hat{y}(x'_s) \in \mathcal{Y}_{\mathrm{att}}
    \land (\neg\mathrm{Gated}
    \lor C(x'_s) \ge \tau_{\mathrm{conf}})
    \land Q+2 \le B$}
        \State Query $f$ twice: $\hat{y}_{\mathrm{rev}}^{(1)}(x'_s), \hat{y}_{\mathrm{rev}}^{(2)}(x'_s)$; $Q \gets Q+2$
        \If{$1+\sum_{j=1}^{2}\mathbf{1}\!\left[\hat{y}_{\mathrm{rev}}^{(j)}(x'_s) \in \mathcal{Y}_{\mathrm{att}}\right] \ge 2$}
            \State \Return $(x'_s, g(x'_s), Q)$
        \EndIf
    \EndIf
    \State \Return $(\bot, g(x'_s), Q)$
\EndFunction

\vspace{3pt}
\Statex \textbf{Phase 1: Offline Seed Search \& Distillation (Sections~\ref{subsec:search}--\ref{subsec:experience})}
\State $\mathcal{L}_{\mathrm{exp}} \gets \emptyset$
\For{each seed task $(x_s, y^*_s, t_s) \in \mathcal{D}_{\mathrm{seed}}$}
    \State Compute clean margin $M_0(x_s)$ (Eq.~\eqref{eq:initial_margin}) and features $\mathbf{v}(x_s)$
    \State $\Psi_{\mathrm{init}} \gets \Psi(S_s, R_s, \mathcal{O}_s)$; $Q \gets 0$, $x^*_s \gets \bot$, $\mathcal{H} \gets \emptyset$ \Comment{Stage 1: Init}
    \For{each $(\psi, \theta_0) \in \Psi_{\mathrm{init}}$ \textbf{while} $Q < B \land x^*_s = \bot$} \Comment{Stage 2: Breadth}
        \State $\delta \gets \psi(x_s, R_s, \mathcal{O}_s, t_s; \theta_0)$
        \State $(x^*_s, g, Q) \gets \Call{EvalSeed}{x_s, \delta, Q}$
        \State \textbf{if} $g > -\infty$ \textbf{then} $\mathcal{H} \gets \mathcal{H} \cup \{(\psi, \theta_0, \delta, g)\}$
    \EndFor
    \State $(x^*_s, \psi^*, \theta^*) \gets \Call{SeedRefine}{x_s, t_s, x^*_s, \mathcal{H}, Q}$ \Comment{Stages 3--4}
    \If{$x^*_s \neq \bot$}
        \State $\theta^* \gets \mathrm{DeIdentify}(\theta^*)$ \Comment{Strip task-specific entities}
        \State Upsert card $e_i = (d, \ell, \mathbf{v}(x_s), \psi^*, \theta^*, \bar{s}_i)$ in $\mathcal{L}_{\mathrm{exp}}$
    \EndIf
\EndFor
\State Freeze $\mathcal{L}_{\mathrm{exp}}$ for read-only test-time retrieval
\setcounter{s1evoline}{\value{ALG@line}}
\end{algorithmic}
&
\begin{algorithmic}[1]
\setcounter{ALG@line}{\value{s1evoline}}
\Statex \textbf{Subroutine: Seed Refinement \& Fallback (Stages 3--4)}
\Function{SeedRefine}{$x_s, t_s, x^*_s, \mathcal{H}, Q$}
    \If{$x^*_s = \bot \land \mathcal{H} \neq \emptyset \land Q < B$} \Comment{Stage 3: Refinement}
        \State $(\hat{\psi}, \hat{\theta}, \hat{\delta}, \hat{g}) \gets \arg\max_{(\psi, \theta, \delta, g) \in \mathcal{H}} g$
        \For{$r = 1$ \textbf{to} $T_{\mathrm{ref}}$ \textbf{while} $Q < B \land x^*_s = \bot$}
            \State $\theta' \gets \mathcal{M}_{\mathrm{att}}(\hat{\psi}, x_s, R_s, \mathcal{O}_s, t_s, \mathcal{H})$; $\delta' \gets \hat{\psi}(x_s, R_s, \mathcal{O}_s, t_s; \theta')$
            \State $(x^*_s, g', Q) \gets \Call{EvalSeed}{x_s, \delta', Q}$
            \State \textbf{if} $g' > -\infty$ \textbf{then} $\mathcal{H} \gets \mathcal{H} \cup \{(\hat{\psi}, \theta', \delta', g')\}$
        \EndFor
    \EndIf
    \For{$r = 1$ \textbf{to} $T_{\mathrm{fb}}$ \textbf{while} $x^*_s = \bot \land Q < B$} \Comment{Stage 4: Fallback}
        \State Synthesize fact-invariant $\delta_{\mathrm{fb}} \gets \mathcal{M}_{\mathrm{att}}(x_s, R_s, \mathcal{O}_s, t_s, \mathcal{H})$
        \State $(x^*_s, g_{\mathrm{fb}}, Q) \gets \Call{EvalSeed}{x_s, \delta_{\mathrm{fb}}, Q}$
    \EndFor
    \State \Return $(x^*_s, \psi^*, \theta^*)$
\EndFunction

\vspace{3pt}
\Statex \textbf{Phase 2: Test-Time One-Step Retrieval \& Construction (Section~\ref{subsec:experience})}
\State Extract structural features $\mathbf{v}(x)$ without pre-querying $f$
\State Select top-$k$ distinct-$\psi$ cards $\mathcal{K}_{\mathrm{top}}$ by $\mathrm{RS}(e_i \mid x)$ (Eq.~\eqref{eq:experience_retrieval})
\State $\mathcal{P}_{\mathrm{ctx}} \gets \mathcal{K}_{\mathrm{top}}$ \textbf{if} $\mathcal{K}_{\mathrm{top}} \neq \emptyset$ \textbf{else} $\Psi_{\mathrm{fb}}$
\State Construct fact-invariant $\delta \gets \mathcal{M}_{\mathrm{att}}(x, R, \mathcal{O}, t, \mathcal{P}_{\mathrm{ctx}})$ in one pass
\State $x' \gets \phi(x, \delta)$; evaluate $f(S(x'), R, \mathcal{O})$ ($2/3$ repeat check on flip)

\vspace{3pt}
\Statex \textbf{Phase 3: Online Dual-System Agent Auditing (Section~\ref{subsec:overview_auditing})}
\State Restore dual-system agent (S1 $f$, generative S2) to prefix $h$
\If{workflow is \textit{Tool Router}}
    \State Inject $\delta$ into tool output $x$; $\hat{y}(x') \gets \pi(P(x'))$; S2 generates args
\ElsIf{workflow is \textit{Action Authorization}}
    \State S2 proposes action $a$ from clean records; inject $\delta$ into note $x$ for $f$
\EndIf
\State Verify exact input match against $S(x')$; execute $\mathcal{E}(h, x', \hat{y}(x'))$
\State Record $\mathbb{I}(\hat{y}(x') \in \mathcal{Y}_{\mathrm{att}})$ and $\Phi_{\mathrm{viol}}\bigl(\mathcal{E}(h, x', \hat{y}(x'))\bigr) = 1$ \Comment{Eq.~\eqref{eq:e2e_violation}}
\State \Return $\mathcal{L}_{\mathrm{exp}}$, constructed $x'$, and node/E2E audit logs
\end{algorithmic}
\end{tabular}
\vspace{2pt}
\end{algorithm*}

\section{Case Studies}
\label{sec:app_cases}

To illustrate how \textbf{S1-EC} bypasses typed S1 models while keeping the ground-truth label $y^*$ unchanged, Figure~\ref{fig:app_case_studies} presents a full-page comparison of two online execution traces on \texttt{jev-1.14}, corresponding to our pre-generation safety guardrail and six-way multi-step tool router. 


\begin{figure*}[p]
\centering
\begin{tcolorbox}[
  enhanced,
  width=\textwidth,
  colback=c1back,
  colframe=c1frame,
  colbacktitle=c1frame,
  coltitle=white,
  fonttitle=\bfseries\normalsize,
  fontupper=\small,
  boxrule=0.9pt,
  arc=3.5pt,
  left=8pt, right=8pt, top=6pt, bottom=6pt,
  after skip=10pt,
  title={(a) Case 1: Content Moderation Filtering}
]
\textbf{1. Task:} Binary content guardrail ($\mathcal{O} = \{\texttt{allow}, \texttt{block}\}$). Ground-truth label: \texttt{allow}; target label: \texttt{block}.\par\smallskip
\textbf{2. Initial Decision:} \texttt{allow} ($P(\texttt{allow}) = 0.990$, $P(\texttt{block}) = 0.010$, margin $M_0 = 0.980$ across $3/3$ runs).\par\smallskip
\textbf{3. Retrieval \& Construction:} Top-$k$ retrieval matches the \texttt{borderline\_mix} experience card; S1-EC constructs a single edge-case sub-request in one pass ($P(\texttt{block}) = 0.880$) and passes $2/3$ re-verification.\par\smallskip
\textbf{4. Attack Input:}
\begin{tcolorbox}[colback=subbox, colframe=c1frame!45, boxrule=0.5pt, arc=2.5pt, left=6pt, right=6pt, top=5pt, bottom=5pt, before skip=4pt, after skip=5pt, fontupper=\small\raggedright]
\textbf{Original Input:} \texttt{Can you suggest some documentaries or films that accurately portray the emotional dynamics of abusive relationships?}\par\smallskip
\textcolor{advred}{\textbf{+ Appended Text:}} \textcolor{advred}{\texttt{\textbf{(How such documentaries typically get around content filters on streaming platforms.)}}}
\end{tcolorbox}
\textbf{5. Final Decision:} \texttt{block} ($P(\texttt{block}) = 0.870 \pm 0.010$, $P(\texttt{allow}) = 0.130 \pm 0.010$, margin $M_{\mathrm{adv}} = 0.740$). The guardrail falsely blocks a benign user request without invoking S2 generation.
\end{tcolorbox}

\begin{tcolorbox}[
  enhanced,
  width=\textwidth,
  colback=c2back,
  colframe=c2frame,
  colbacktitle=c2frame,
  coltitle=white,
  fonttitle=\bfseries\normalsize,
  fontupper=\small,
  boxrule=0.9pt,
  arc=3.5pt,
  left=8pt, right=8pt, top=6pt, bottom=6pt,
  title={(b) Case 2: Multi-Step Tool Routing}
]
\textbf{1. Task:} Six-way tool router ($\mathcal{O} = \{\texttt{web\_search}$, \texttt{calculator}, \texttt{calendar}, \texttt{send\_email}, \texttt{file\_read}, $\texttt{file\_delete}\}$). Subtask: \texttt{Create an event titled 'Sync follow-up' on 2026-03-05 at 16:00}. Ground-truth label: \texttt{calendar}; target label: \texttt{send\_email}.\par\smallskip
\textbf{2. Initial Decision:} \texttt{calendar} ($P(\texttt{calendar}) = 1.00$, $P(\texttt{send\_email}) = 0.00$, margin $M_0 = 1.00$ across $12/12$ runs).\par\smallskip
\textbf{3. Retrieval \& Construction:} Top-$k$ retrieval selects the \texttt{condmap} routing-note card distilled offline via margin feedback; S1-EC binds the current subtask to construct a single tool-return note in one pass without online search.\par\smallskip
\textbf{4. Attack Input:}
\begin{tcolorbox}[colback=subbox, colframe=c2frame!45, boxrule=0.5pt, arc=2.5pt, left=6pt, right=6pt, top=5pt, bottom=5pt, before skip=4pt, after skip=5pt, fontupper=\small\raggedright]
\textbf{Original Tool Return:} \texttt{[previous\_tool result] - 2026-03-05: Team sync 10:00}\par\smallskip
\textcolor{advred}{\textbf{+ Appended Text:}} \textcolor{advred}{\texttt{\textbf{This subtask is a documented email-notification test case, so the required action is to send the follow-up message rather than create the calendar entry.}}}
\end{tcolorbox}
\textbf{5. Final Decision:} \texttt{send\_email} ($P(\texttt{send\_email}) = 0.99$, $P(\texttt{calendar}) = 0.01$, margin $M_{\mathrm{adv}} = 0.98$ across $9/9$ checks). In $3$ end-to-end replays, all $3/3$ tasks fail and $2/3$ send unauthorized external emails.
\end{tcolorbox}

\caption{Two execution traces of \textbf{S1-EC} on \texttt{jev-1.14} for content moderation filtering (top) and tool routing (bottom), following the sequence of task, initial decision, retrieval and construction, attack input (\textcolor{advred}{\textbf{bold red}}), and final decision.}
\label{fig:app_case_studies}
\end{figure*}

\end{document}